\documentclass[letterpaper, 10 pt, journal, twoside]{IEEEtran}
\usepackage{amsmath,amsfonts}
\usepackage{algorithmic}
\usepackage{array}
\usepackage[caption=false,font=normalsize,labelfont=sf,textfont=sf]{subfig}
\usepackage{caption}
\usepackage{textcomp}
\usepackage{stfloats}
\usepackage{url}
\usepackage{verbatim}
\usepackage{graphicx}
\usepackage[table]{xcolor}
\usepackage{booktabs}
\usepackage{cite}
\usepackage{microtype}
\usepackage{bm}
\usepackage{url}

\definecolor{bestcell}{HTML}{DCE7F5}
\providecommand{\bc}{\cellcolor{bestcell}}
\newcommand{\vsep}{\multicolumn{1}{@{\hspace{6pt}}c@{\hspace{6pt}}|@{\hspace{6pt}}}{}}
\newcommand{\vsepn}{\multicolumn{1}{@{\hspace{3pt}}c@{\hspace{3pt}}|@{\hspace{3pt}}}{}}
\ifdefined
\fi

\begin{document}
\title{CoDDP: Coordinated Decentralized Diffusion Policies with Semantic-Kinematic Affordances for Bimanual Manipulation}
\author{Ningze Sun and Roya Firoozi%
\thanks{N. Sun and R. Firoozi are with the University of Waterloo, Waterloo, ON, Canada (e-mail: n7sun@uwaterloo.ca; roya.firoozi@uwaterloo.ca).}}
\maketitle

\begin{abstract}
Bimanual manipulation requires two arms to divide roles over time, track task progress, and ground actions in a shared scene. Policies that map a joint observation to a joint action couple the two arms inside a single network, so neither agent can be perturbed or delayed on its own. We present CoDDP, a bimanual diffusion policy under centralized training with decentralized execution. The two arms are separate diffusion policies acting on their own observations, coupled at execution through a communication channel. We introduce two modules that make this decomposition work: a semantic-kinematic affordance encoder that predicts where each arm and its partner should act, and a phase-based 3D trace that guides an arm's spatial motion. On selected RoboTwin~2.0 tasks, CoDDP is competitive with fine-tuned monolithic vision-language-action baselines while remaining robust when one agent's sensing is corrupted or its motion is delayed. Ablations and a mutual-information analysis further characterize the contributions of the proposed components and communication channel. Project page: \url{https://ningzesun.github.io/coddp/}.
\end{abstract}


\section{Introduction}

Many manipulation tasks require two arms to act and coordinate within the same scene, from handing an object between grippers to stabilizing a pot while filling it. Recent imitation-learned policies handle such tasks by treating the dual-arm system as a single agent, in which one network takes joint observations as input and outputs joint actions~\cite{zhao2023act,black2024pi0,liu2025rdt,intelligence2025pi05,dm05}. Although effective, this monolithic design tightly couples the two arms through a shared policy, limiting their ability to adapt independently to local disturbances. For example, a sensor fault affecting one arm can compromise the policy's action generation for both arms. Moreover, an execution delay in one arm can disrupt the coordination of the entire system.

Multi-agent reinforcement learning offers an alternative through centralized training with decentralized execution (CTDE)~\cite{lowe2017maddpg,rashid2018qmix,yu2022mappo}, where agents are trained jointly with access to global information but execute their policies using local observations, optionally exchanging explicit messages. Applied to bimanual manipulation, this decomposition separates the policy execution of the two arms, allowing each to operate independently of its partner's sensing and execution pipeline. Consequently, a sensor fault or state corruption affecting one arm need not directly compromise the other arm's action generation. Similarly, an execution delay in one arm need not interrupt the other's progress. Such resilience is particularly important in manufacturing settings, where a localized sensor failure, corrupted state estimate, or delayed part arrival should not bring an entire assembly station to a halt. Yet naive decentralization discards exactly what bimanual tasks need most: each arm must still cooperate with its partner, infer its intentions, and determine where it should act next.

To address this challenge we present CoDDP, a bimanual decentralized diffusion policy that decomposes coordinated manipulation into two complementary forms of grounding: (i) what and whom to interact with, (ii) and how to move. First, a semantic-kinematic affordance encoder (SKAE) grounds each arm's actions in the shared scene by identifying the object parts to interact with and assigning interaction roles to the two arms. Using frozen visual-language features, it predicts both its own and its partner's affordances and identifies the arm associated with each affordance. The affordances are \emph{semantic} because they identify the relevant object part to contact from the scene and instruction, and \emph{kinematic} because they also depend on the arm's state and its partner's gripper closure, allowing the interaction grounding to track each arm's next contact as the task progresses. Second, a phase-based 3D trace adapted from~\cite{robobrain25,liu2026lohomanip} then converts this interaction grounding into spatial motion guidance by predicting the arm's upcoming tool-center path (the sequence of 3D positions that the robot's end effector is expected to follow over time) as pixel coordinates with depth. Thus, SKAE establishes which object parts each arm should interact with and assigns the corresponding interaction roles, while the 3D trace predicts how each arm intends to realize these interactions through spatial motion and shares this motion intent with its partner through communication. Finally, each arm runs its own diffusion policy, which denoises an action chunk conditioned on four inputs: the arm's local observation, its predicted own and partner affordances, its predicted trace, and the received partner message. Together, these grounding modules enable coordinated bimanual execution through decentralized action generation, while accommodating communication delays and failures.

We evaluate CoDDP on held-out out-of-distribution (OOD) scenes of selected bimanual tasks of RoboTwin~2.0~\cite{chen2025robotwin2}, comparing it against a decentralized diffusion baseline, monolithic vision-language-action (VLA) baselines, and architectural ablations. Beyond task success, we evaluate the properties that motivate the decentralized design. Specifically, we inject sensor faults and action delays into one agent to assess policy robustness. To show that each agent relies on the communication channel to act, we directly quantify the information conveyed through the channel using a variational estimate of the conditional mutual information between an agent’s action and the message it receives.

In summary, we make the following contributions.

\begin{itemize}
  \item \textbf{Decentralized bimanual framework (CoDDP)}: We introduce CoDDP with two arm-specific diffusion policies that execute from local observations and coordinate through a compact communication channel. CoDDP integrates affordance grounding and phase-conditioned 3D motion guidance to enable coordinated manipulation under decentralized execution, including single-arm sensing faults and execution delays.
  \item \textbf{Parallel affordance grounding (SKAE)}: We develop a semantic-kinematic affordance encoder that jointly predicts each arm's own and partner's affordances, using an agent-specific-affordance (ASA) score for role assignment. This provides each arm with explicit coordination cues without requiring centralized observations.
  \item \textbf{Phase-conditioned motion guidance (3D trace)}: We introduce short-horizon, depth-aware 3D tool-center-point traces, predicted from affordance-supervised tokens. These traces condition each arm's own policy and are sent to the partner through the communication channel, conveying the arm's motion intent for inter-arm coordination.
  \item \textbf{Comprehensive Evaluation}: We evaluate CoDDP on 15 RoboTwin~2.0 tasks, demonstrating performance comparable to fine-tuned monolithic VLAs and improved robustness to single-arm sensing faults and execution delays. Ablations and mutual-information analysis further characterize the contributions of the proposed components and communication channel.
\end{itemize}

\section{Related Work}

\subsection{Imitation Learning for Manipulation}

Recent manipulation policies learn continuous control directly from demonstrations. Diffusion policies~\cite{chi2023diffusion} model the action distribution through an iterative denoising process, while action-chunking transformers~\cite{zhao2023act} predict short action sequences to reduce compounding errors. RDT-1B~\cite{liu2025rdt} scales the diffusion transformers~\cite{peebles2023dit} into a large bimanual foundation model. VLA such as $\pi_0$~\cite{black2024pi0}, $\pi_{0.5}$~\cite{intelligence2025pi05}, and DM0.5~\cite{dm05} extend this paradigm by integrating action generation with pretrained vision-language representations.

These methods treat the two arms as a single agent with a joint observation and a joint action space. Consequently, both arms' actions are generated through a centralized common policy and conditioned on shared information. This centralized coupling can make the policy sensitive to sensing failures affecting either arm and limits the ability to independently adapt each arm's behavior to local observations or execution errors. CoDDP retains diffusion-based action generation and action chunking but decomposes the centralized policy into two coordinated, decentralized policies.

\subsection{Multi-Agent Learning and Communication}

Centralized training with decentralized execution (CTDE) is a widely used paradigm for learning agents that execute using local observations. MADDPG~\cite{lowe2017maddpg}, QMIX~\cite{rashid2018qmix}, and MAPPO~\cite{yu2022mappo} use centralized information during training while maintaining decentralized execution. A related line of work investigates how agents can communicate, including CommNet~\cite{sukhbaatar2016commnet} and TarMAC~\cite{das2019tarmac}, which jointly learn communication mechanisms and policies. These approaches primarily study reinforcement learning in low-dimensional state-based environments rather than vision-based manipulation.

More closely related to our work are MIMIC-D~\cite{dong2025mimicd} and LatentToM~\cite{he2025latenttom}, which learn decentralized diffusion policies for cooperative manipulation from multi-agent demonstrations. MIMIC-D jointly trains agent-specific diffusion policies using local observations and a shared coordination loss, enabling implicit coordination during decentralized execution. LatentToM learns ego and consensus embeddings that capture relationships between agents, enabling coordination through shared visual observations or explicit exchange of latent representations. Neither approach integrates language-conditioned task instructions or explicitly investigates robustness to scene-level distribution shifts. CoDDP extends decentralized diffusion-based imitation learning with visual-language grounding and an explicit inter-arm communication channel, whose contribution we systematically evaluate through ablations and robustness experiments.

CHORUS~\cite{doshi2026chorus} deploys independent instances of a shared VLA policy on each robot, relying on visual observations for coordination without explicit communication at inference. In contrast, CoDDP's communication channel allows each arm to access information about its partner's progress, enabling coordinated bimanual manipulation under decentralized execution, even when one arm experiences execution delays.

\subsection{Affordance and Spatial Guidance}

Rather than learning where to act entirely end to end, many manipulation systems predict affordances, the regions where object interaction should occur~\cite{mo2021where2act}, and recent work incorporates such predictions into a VLA~\cite{yu2026affordancevla}. Open-vocabulary predictors ground a language query to one agent's interaction region~\cite{shorinwa2024splatmover,yuan2024robopoint}, and DualAfford~\cite{zhao2023dualafford} predicts a second gripper's affordance conditioned on the first gripper's selected action. None conditions on the arms' states, so the next contact comes from the query not from task progress, and DualAfford's sequential prediction limits parallel decision-making. We therefore train our own predictor, in which each arm, in parallel, predicts its own and its partner's next contact from its local observation, state, and partner's gripper closure.

A complementary form of spatial guidance is a trace: an ordered sequence of future keypoints that guides gripper motion. RoboBrain~2.5~\cite{robobrain25} predicts depth-aware 3D manipulation traces, while LoHo-Manip~\cite{liu2026lohomanip} uses 2D traces to guide downstream execution. Unlike these externally generated planning cues, CoDDP incorporates phase-conditioned traces directly into the policy, predicting short-horizon keypoints from affordance-supervised tokens.

\begin{figure*}[t]
  \centering
  \includegraphics[width=\textwidth]{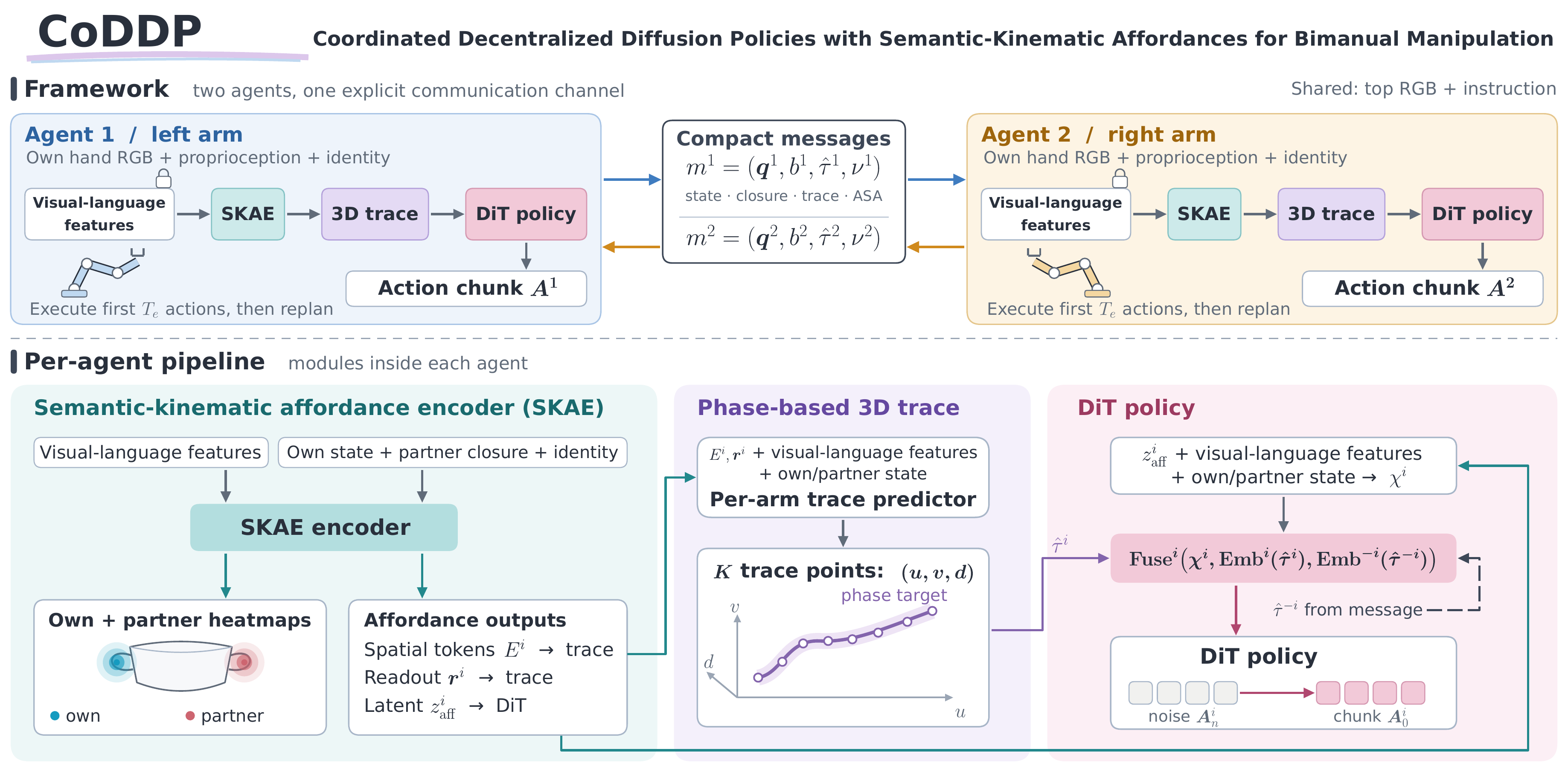}
  \caption{\textbf{CoDDP overview.}
  Top: framework. Two decentralized per-arm policies coordinate through compact messages while producing separate action chunks. Bottom: per-agent pipeline. SKAE (left) predicts own and partner affordances and feeds its latent and spatial tokens downstream; the phase-based trace (middle) predicts the trajectory to the current motion-phase target; and the DiT policy (right) fuses visual-language, state, affordance, and own and partner trace conditions to denoise each arm's action chunk.}
  \label{fig:overview}
\end{figure*}

\section{Problem Formulation}

In this problem, two robotic manipulator arms must cooperate to complete a task specified by a natural language instruction by learning from expert demonstrations. Although each arm controls its own joints and end effector, successful execution requires implicit coordination: the arms must avoid collisions with one another, assign roles over time, and ground their actions in the visual scene.

The system consists of two agents $\mathcal{I}=\{1,2\}$, corresponding to the left and right manipulator arms; we write $-i$ for the partner of agent $i$. At each time step, agent $i$ outputs an action $a^i_t\in\mathcal{A}^i$ in its action space $\mathcal{A}^i$, which contains the end-effector-space target as well as one gripper command.
The raw observation available to agent $i$ at time $t$ is $\bar{o}^i_t=(X^{top}_t,X^{hand,i}_t,\bm{q}^i_t,\ell,i)$, where $X^{top}_t$ is the shared top RGB camera image, $X^{hand,i}_t$ is the hand-camera image for arm $i$, $\bm{q}^i_t$ is the proprioceptive state of arm $i$, $\ell$ is the natural language instruction, and $i$ is the agent identity.
The dataset $\mathcal{D}=\{(\xi^{1,(j)},\xi^{2,(j)},\bar{o}^{(j)},\ell^{(j)})\}_{j=1}^{M}$ consists of $M$ demonstrations, where $\xi^{i,(j)}=\{a^i_0,\ldots,a^i_{T-1}\}$ is the expert action trajectory of arm $i$ over $T$ steps, and $\bar{o}^{(j)}$ and $\ell^{(j)}$ are the recorded raw observations and the instruction.

From $\mathcal{D}$, we aim to learn a decentralized policy $\pi^i$ for each agent that acts on its own observation $\bar{o}^i_t$ and a message received from its partner, instead of a single policy that maps the joint observation to the joint action. Separating the agents keeps each arm's action generation independent of its partner's sensing and execution pipeline, so a sensor fault or an execution delay in one arm need not compromise the other.

\emph{Notation.} Superscript $i$ indexes the agent and subscript $t$ the time step. Unless defined otherwise, a bar marks a raw quantity, a hat a predicted quantity (e.g., the predicted trace $\hat{\tau}^i_t$ versus its label $\tau^i_t$), and a tilde a state augmented with the gripper-closure bit, $\tilde{\bm{q}}^i_t=[\bm{q}^i_t,b^i_t]$, where $b^i_t\in\{0,1\}$ is taken from demonstration labels during training and from proprioception at execution. Frozen pretrained vision and vision-language encoders (Appendix) provide the visual-language features: lowercase $f^{top}_t$ and $f^{hand,i}_t$ are the global top-camera and hand-camera features, uppercase $F^{top}_t$ and $F^{hand,i}_t$ are the corresponding patch-token grids of $G\times G$ tokens that retain spatial layout, and $f^{\ell}$ encodes the instruction $\ell$.

\section{Method}

\subsection{System Overview}

CoDDP consists of five components, summarized in Fig.~\ref{fig:overview}: the top shows the decentralized framework, and the bottom shows the per-agent pipeline. First, frozen visual-language backbones encode top-camera images, hand-camera images, and language instructions. Second, a decentralized DiT diffusion policy predicts action chunks for each arm separately. Third, a compact communication channel carries each arm's state, gripper-closure bit, agent-specific-affordance (ASA) logit, and predicted trace to its partner. Fourth, SKAE produces per-arm affordance latents, affordance heatmaps, affordance-supervised spatial tokens, and the ASA logit. Fifth, a phase-based 3D trace module provides spatial guidance for each arm.

The method separates semantic grounding, affordance prediction, spatial motion guidance, and action generation. This separation is useful for bimanual manipulation because the policy must not only generate smooth actions, but also preserve task progress and coordinate role changes between arms.

\subsection{Decentralized DiT Policy}\label{sec:dit}

From the raw observation, agent $i$ derives a structured observation $o^i_t=(f^{top}_t,f^{hand,i}_t,F^{hand,i}_t,f^{\ell},\tilde{\bm{q}}^i_t,z^i_{\mathrm{aff},t},\hat{\tau}^i_t,i)\in\mathcal{O}^i$ in its observation space $\mathcal{O}^i$, where $z^i_{\mathrm{aff},t}$ is the SKAE affordance latent (Section~\ref{sec:skae}) and $\hat{\tau}^i_t$ is the predicted phase-based trace (Section~\ref{trace}). At time step $t$, each agent selects actions according to a decentralized policy $a^i_t\sim\pi^i_{\theta^i}(a^i_t\mid o^i_t,m^{-i}_t)$ with parameters $\theta^i$, where $m^{-i}_t$ is the partner message (Section~\ref{sec:comm}). Each policy is parameterized as a conditional diffusion model over action chunks of horizon $T_a$. Let $\bm{A}^i_0=a^i_{t:t+T_a-1}$ denote the clean expert action chunk. The forward diffusion process samples $\bm{A}^i_n=\sqrt{\bar{\alpha}_n}\bm{A}^i_0+\sqrt{1-\bar{\alpha}_n}\,\epsilon$ with $\epsilon\sim\mathcal{N}(0,I)$, where $n$ is the diffusion step and $\bar{\alpha}_n$ is the cumulative noise schedule. The policy predicts the injected noise with a denoiser $\epsilon^i_\theta(\bm{A}^i_n,n,\psi^i_t)$, a diffusion transformer~\cite{peebles2023dit} over the $T_a$ action tokens, modulated by the conditioning vector $\psi^i_t$ and the diffusion-step embedding through adaptive layer normalization. The conditioning vector is produced from the current observation and the received partner state, and then fused with the own and partner predicted traces:
\begin{equation*}
\begin{split}
\chi^i_t&=\mathrm{Cond}^{i}\big(z^i_{\mathrm{aff},t},f^{top}_t,f^{hand,i}_t,F^{hand,i}_t,f^{\ell},\tilde{\bm{q}}^i_t,\tilde{\bm{q}}^{-i}_t\big),\\
\psi^i_t&=\mathrm{Fuse}^{i}\big(\chi^i_t,\mathrm{Emb}^{i}(\hat{\tau}^i_t),\mathrm{Emb}^{-i}(\hat{\tau}^{-i}_t)\big),
\end{split}
\end{equation*}
where $\mathrm{Cond}^{i}$ and $\mathrm{Fuse}^{i}$ are learned networks of arm $i$, $\mathrm{Emb}^{i}$ is a trace embedding (Appendix), and the partner state $\tilde{\bm{q}}^{-i}_t$ and trace $\hat{\tau}^{-i}_t$ are received in $m^{-i}_t$. The two policies are trained jointly under CTDE on the summed denoising losses,
\[
\mathcal{L}=\sum_{i\in\mathcal{I}}\mathcal{L}^{i}_{\mathrm{diff}},\quad
\mathcal{L}^{i}_{\mathrm{diff}}=\mathbb{E}_{\mathcal{D},n,\epsilon}\big\|\epsilon-\epsilon^{i}_{\theta}(\bm{A}^i_n,n,\psi^i_t)\big\|_2^2.
\]
Training is centralized only in that the two policies are optimized jointly on this summed loss, with no additional coupling between the agents. Each policy is still trained on partial observations, receiving exactly the inputs it receives at execution. Separate left-arm and right-arm conditioners and denoisers share no parameters, so one agent's policy can be retrained or replaced without affecting the other.

\subsection{Communication Channel}\label{sec:comm}

Each agent broadcasts a compact message $m^i_t=(\bm{q}^i_t,b^i_t,\hat{\tau}^i_t,\nu^i_t)\in\mathcal{M}$ in the message space $\mathcal{M}$, which agent $-i$ receives at the same time step; $\nu^i_t$ is the ASA logit of the SKAE (Section~\ref{sec:skae}). The message enters the partner's computation in three places: its state and closure bit are concatenated into the partner's conditioning state, its ASA logit normalizes the partner's ASA score, and its trace is fused with the partner's own predicted trace before conditioning the denoiser. Removing the channel ($\mathcal{M}=\emptyset$) results in two independent actors that share only the top-camera image and instruction.

\subsection{Semantic-Kinematic Affordance Encoder}\label{sec:skae}

Semantic-Kinematic Affordance Encoder (SKAE) maps visual, language, state, and identity inputs into an arm-specific affordance representation. The affordance is \emph{semantic} because it is read from frozen visual-language features of the scene and instruction and is supervised on the contacted object part. It is \emph{kinematic} because it is conditioned on the arm's state and its partner's gripper-closure bit, and its targets are each arm's next contact, so it changes as the task progresses.

For arm $i$, the encoder input tokens are the visual-language features, the augmented arm state $\tilde{\bm{q}}^i_t$, the partner's gripper-closure bit $b^{-i}_t$, and a learned agent identity embedding $e^i$. Both arms share the SKAE weights, so $e^i$ tells the encoder which arm it is predicting for. These tokens are projected and processed by the encoder:
\[
[h^i_t,E^i_t]=\mathrm{Encoder}\big(f^{top}_t,F^{top}_t,f^{hand,i}_t,f^{\ell},\tilde{\bm{q}}^i_t,b^{-i}_t,e^i\big),
\]
where $h^i_t$ is a global output that summarizes all input tokens, and $E^i_t$ are the outputs at the patch positions of $F^{top}_t$. A projection head $g_{\mathrm{lat}}$ maps $h^i_t$ to the affordance latent $z^i_{\mathrm{aff},t}=g_{\mathrm{lat}}(h^i_t)$. The latent also feeds a scalar agent-specific-affordance (ASA) head that outputs the ASA logit $\nu^i_t=g_{\mathrm{asa}}(z^i_{\mathrm{aff},t})$. With the partner's logit $\nu^{-i}_t$ received through the message, the logits are normalized across the two arms into the ASA score $\hat{s}^i_t=\exp(\nu^i_t)/\sum_{j\in\mathcal{I}}\exp(\nu^j_t)$, predicting which arm the current affordance belongs to. When both arms have affordance targets, the scores split toward $0.5$.

A convolutional decoder upsamples the spatial tokens $E^i_t$ to two heatmaps:
\[
\big(\hat{H}^{i,\mathrm{own}}_t,\,\hat{H}^{i,\mathrm{partner}}_t\big)=\mathrm{Decoder}\big(E^i_t\big).
\]
The own-affordance heatmap predicts where arm $i$ should act. The partner-affordance heatmap predicts where the other arm should act. These heatmaps provide spatial grounding and encourage the latent to encode coordination-relevant scene structure, while $E^i_t$ becomes an affordance-supervised token grid that the trace (Section~\ref{trace}) reuses. Fig.~\ref{fig:qualitative} shows both agents' own and partner heatmaps.

\newlength{\qualpanelw}
\setlength{\qualpanelw}{\dimexpr(\textwidth-62pt)/6\relax}
\newcommand{\nodepth}[1]{\raisebox{\depth}{#1}}
\newcommand{\tasklab}[1]{%
  \nodepth{\parbox[c][0.75\qualpanelw][c]{44pt}{\centering\scriptsize #1}}}
\definecolor{ownaff}{HTML}{00D1FF}
\definecolor{partneraff}{HTML}{FF1F38}
\definecolor{lefttrace}{HTML}{FFD400}
\definecolor{righttrace}{HTML}{B84DFF}

\begin{figure*}[t]
  \centering
  \setlength{\tabcolsep}{1.5pt}
  \renewcommand{\arraystretch}{0.9}
  \begin{tabular}{@{}c@{\hspace{2pt}}cccccc@{}}
     & \scriptsize\shortstack{Third-\\person} & \scriptsize RGB
     & \scriptsize\shortstack{Left\\affordance} & \scriptsize\shortstack{Right\\affordance}
     & \scriptsize\shortstack{Left\\trace} & \scriptsize\shortstack{Right\\trace} \\[1pt]
    \tasklab{Place bread\\on skillet}
     & \includegraphics[width=\qualpanelw]{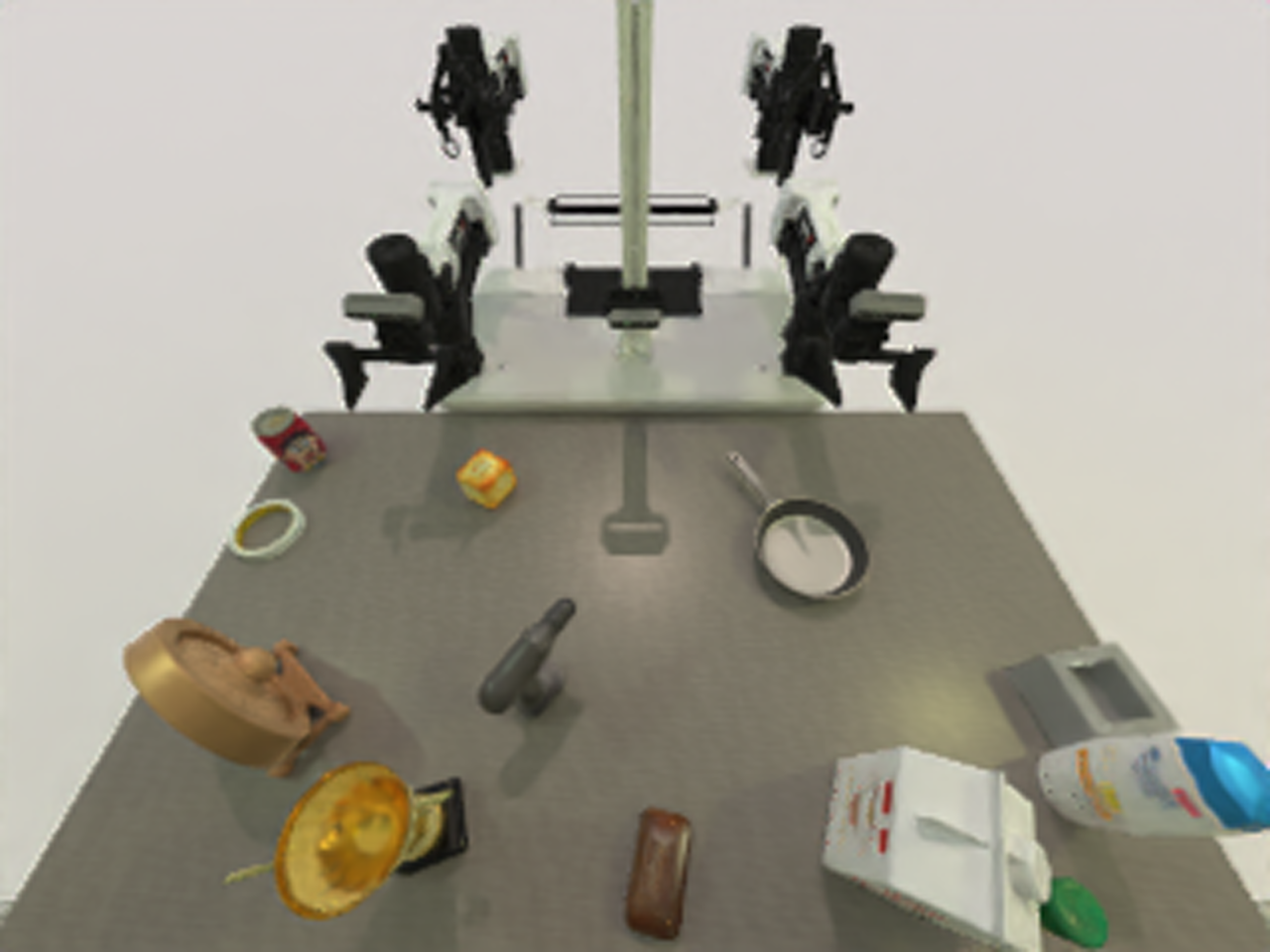}
     & \includegraphics[width=\qualpanelw]{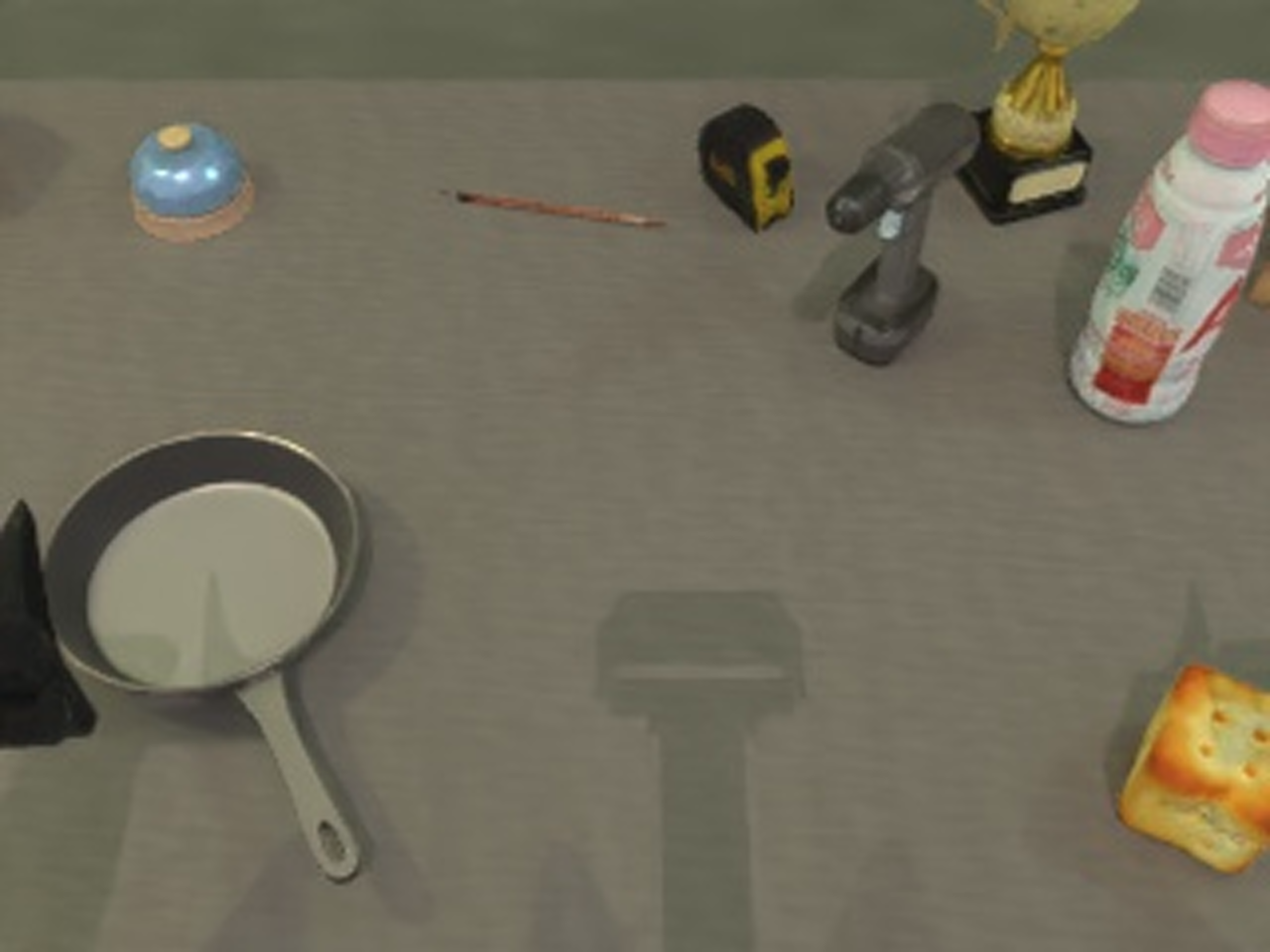}
     & \includegraphics[width=\qualpanelw]{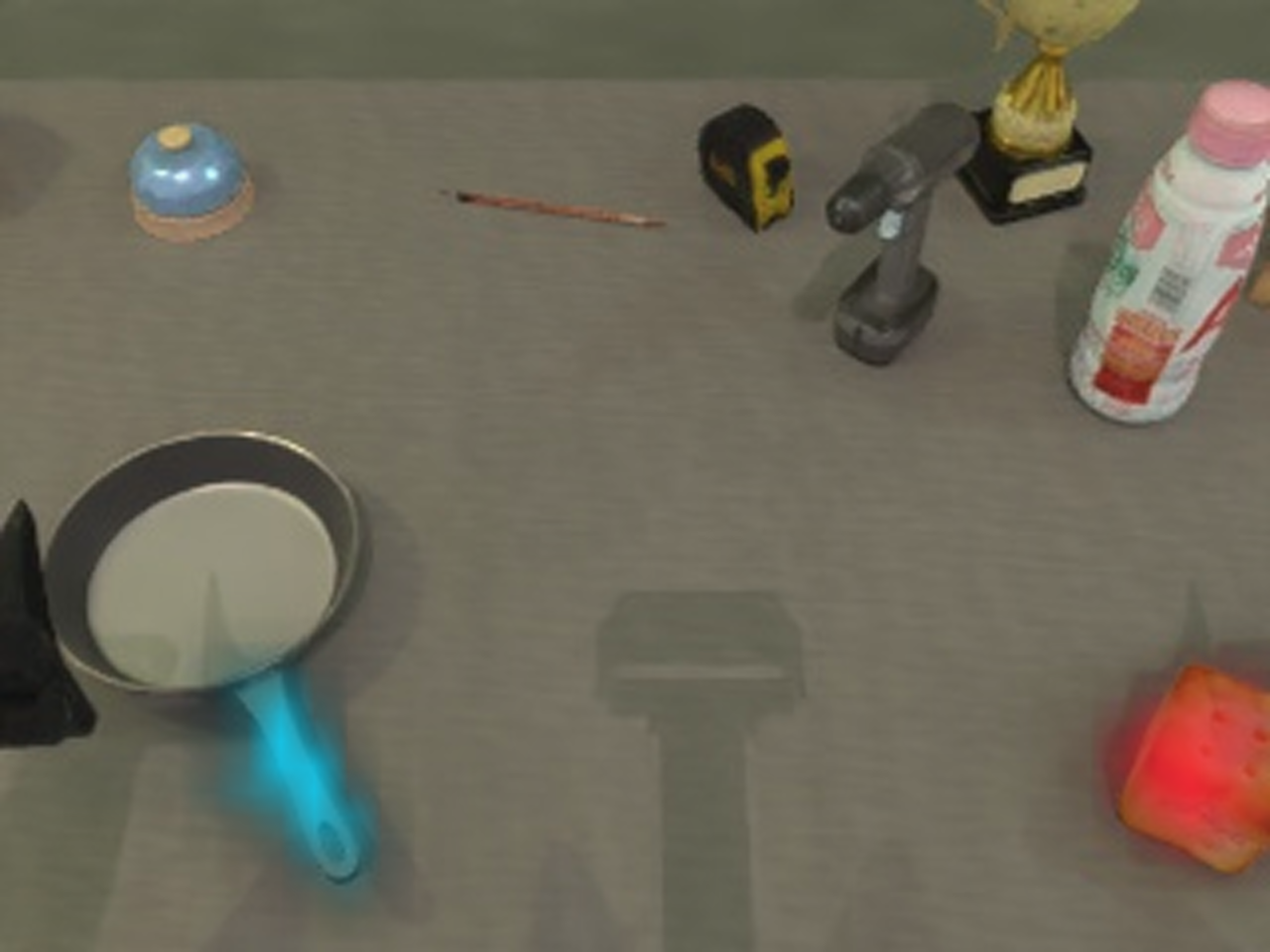}
     & \includegraphics[width=\qualpanelw]{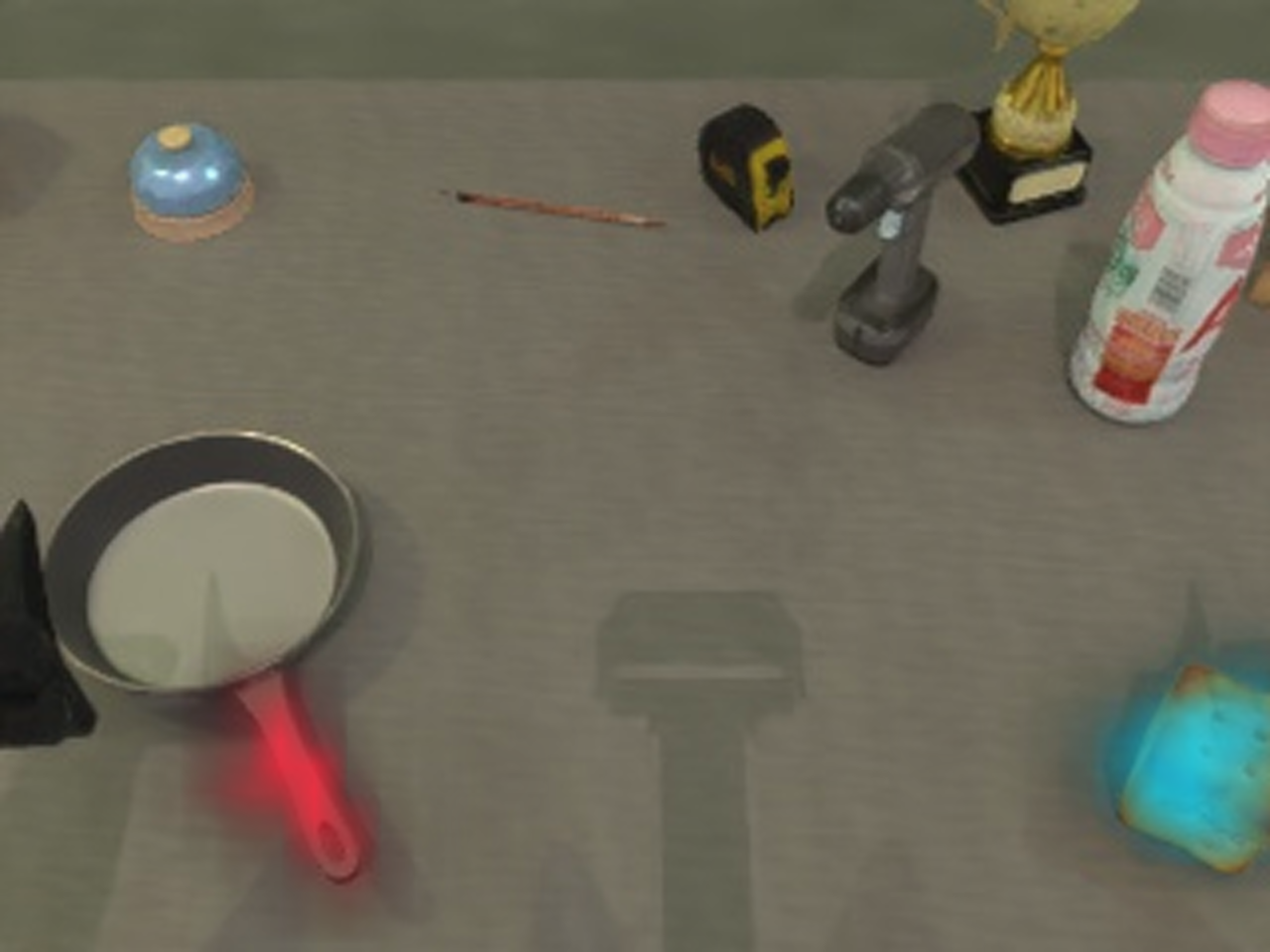}
     & \includegraphics[width=\qualpanelw]{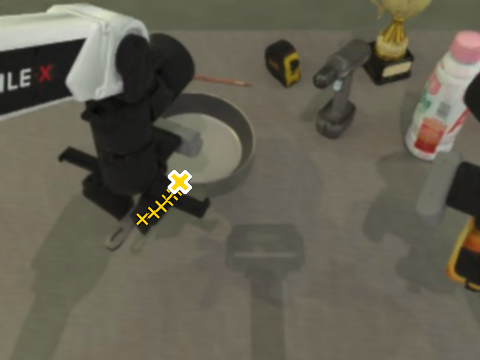}
     & \includegraphics[width=\qualpanelw]{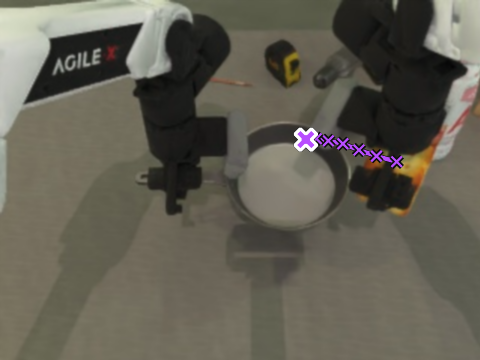} \\[1pt]
    \tasklab{Lift pot}
     & \includegraphics[width=\qualpanelw]{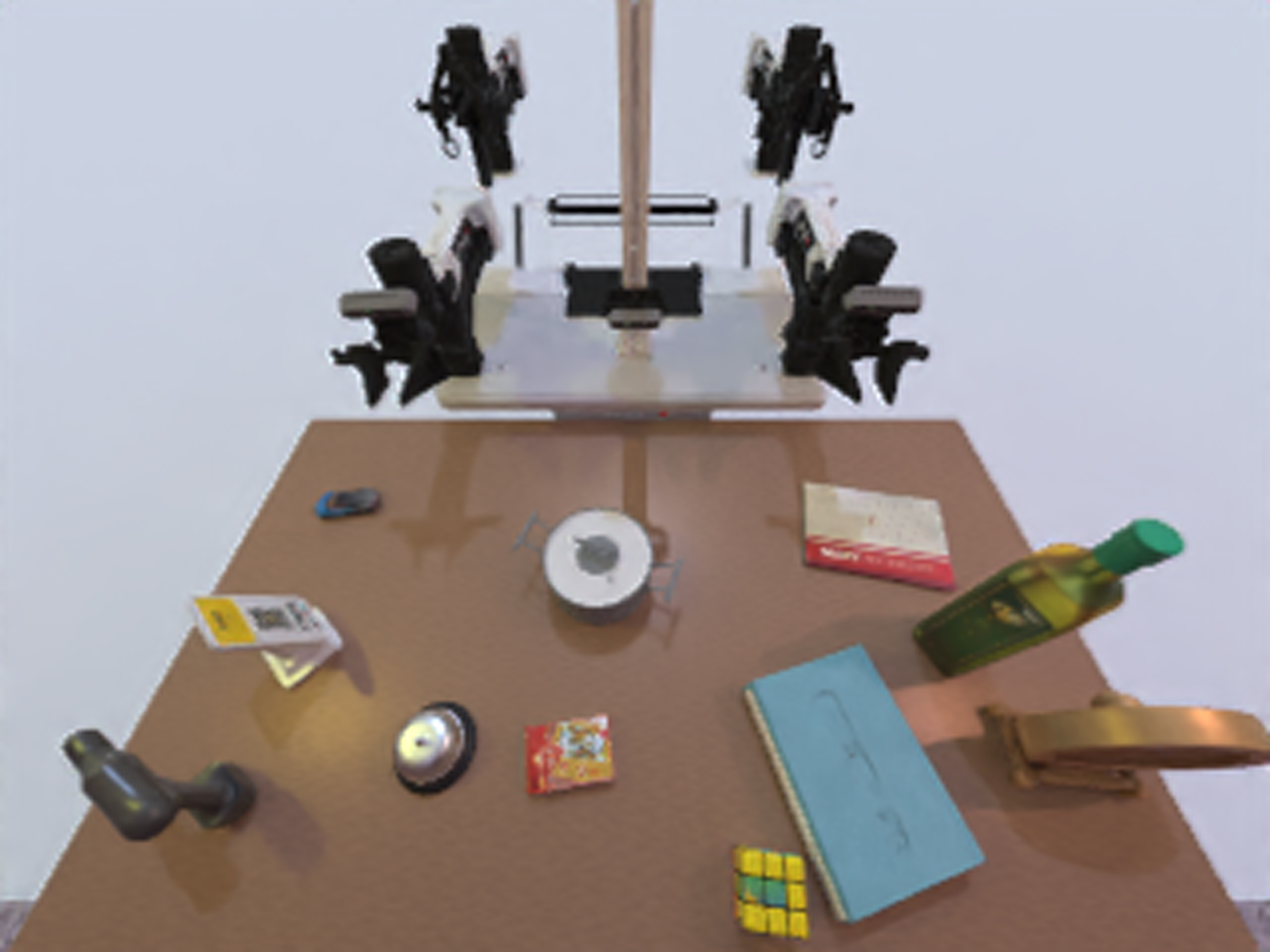}
     & \includegraphics[width=\qualpanelw]{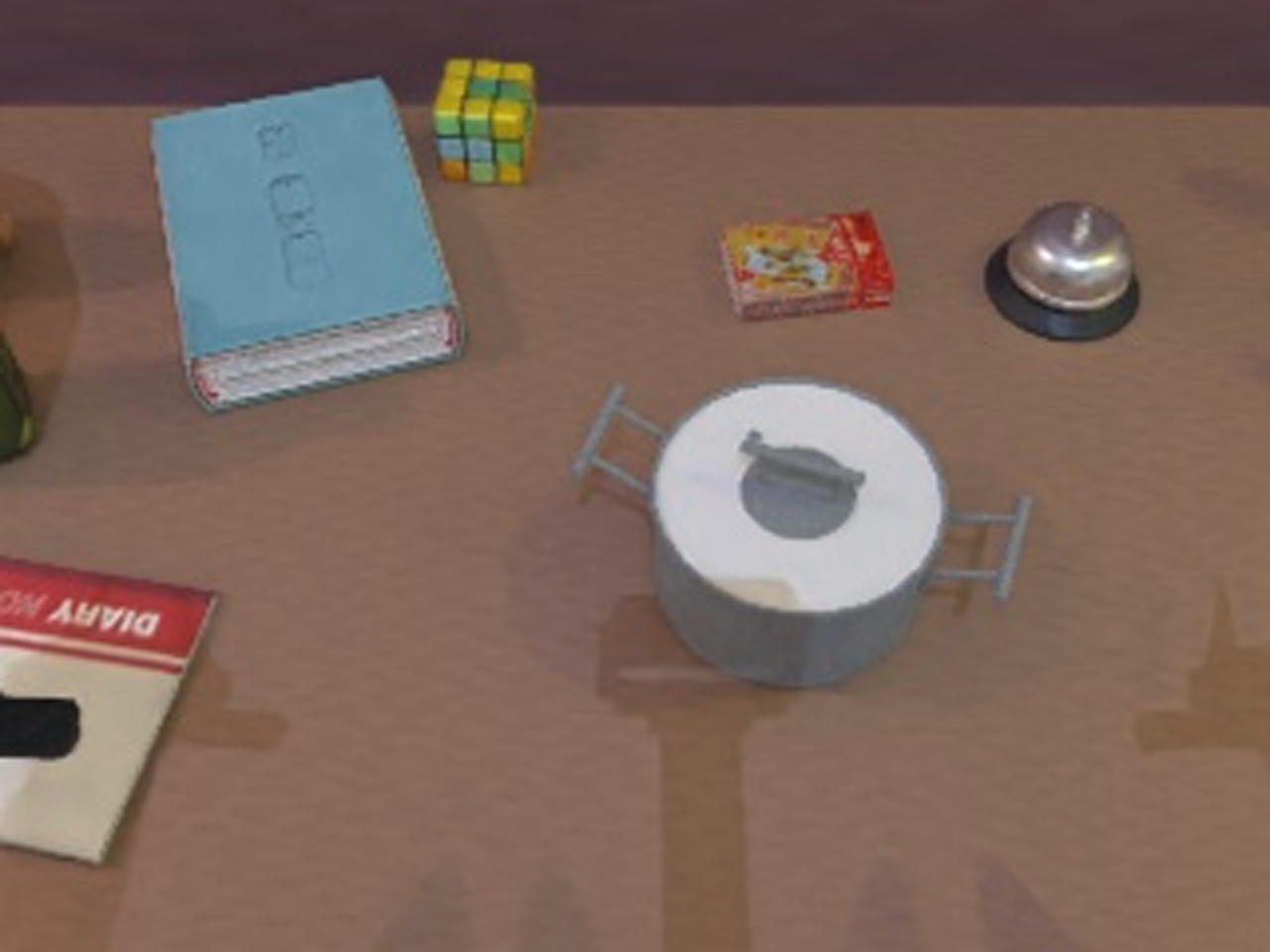}
     & \includegraphics[width=\qualpanelw]{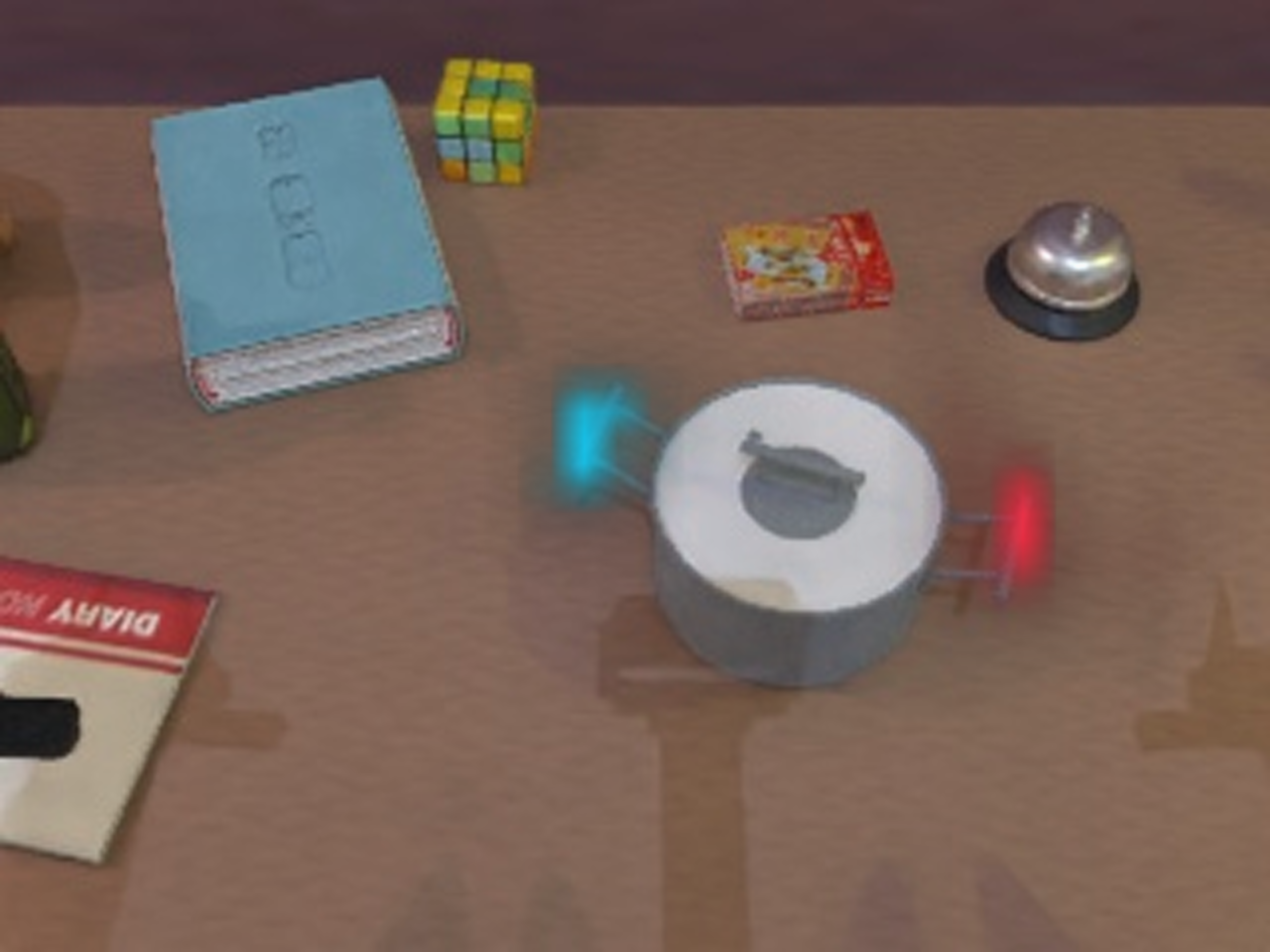}
     & \includegraphics[width=\qualpanelw]{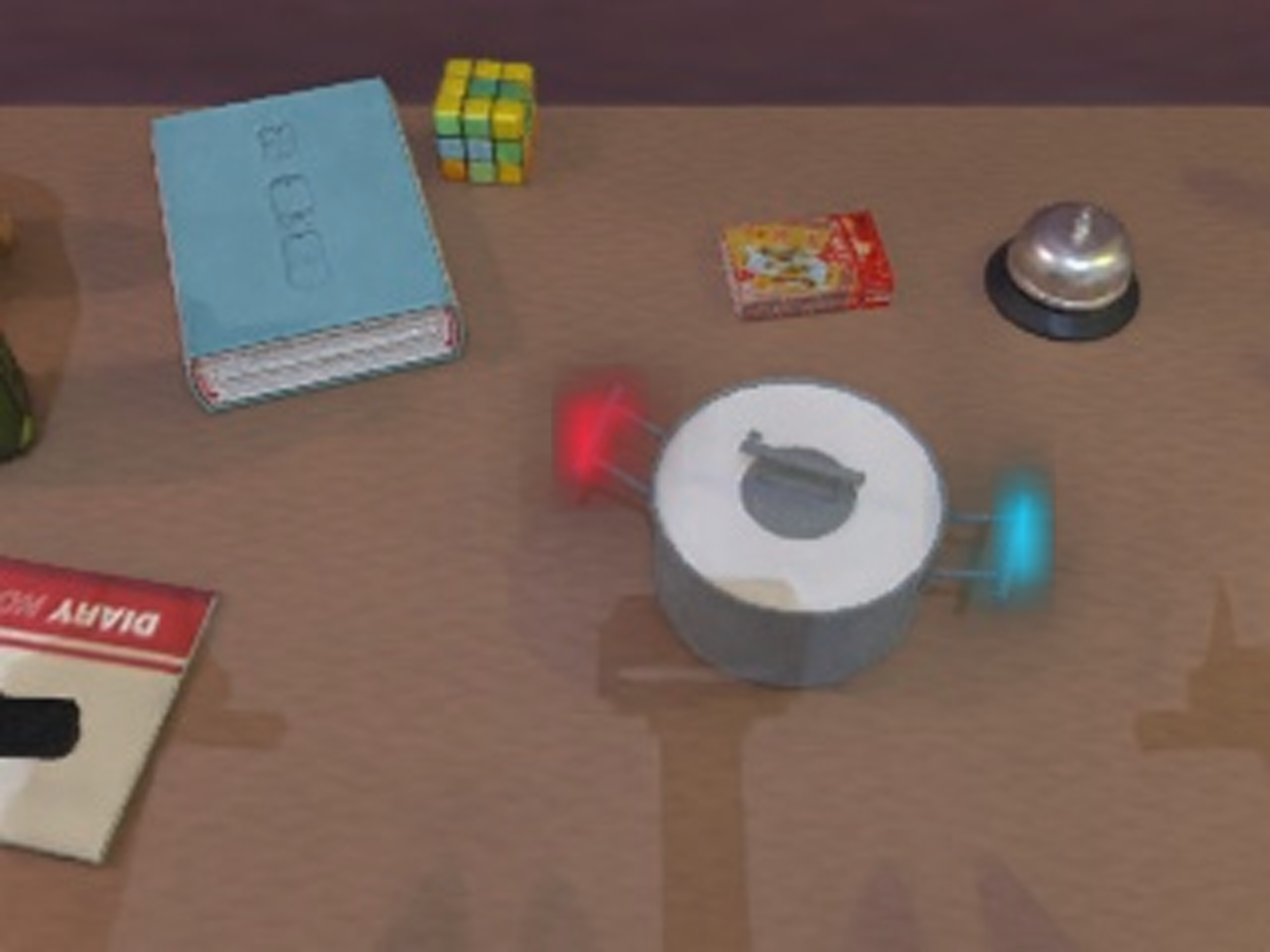}
     & \includegraphics[width=\qualpanelw]{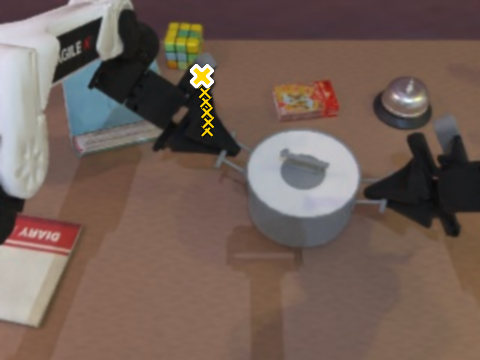}
     & \includegraphics[width=\qualpanelw]{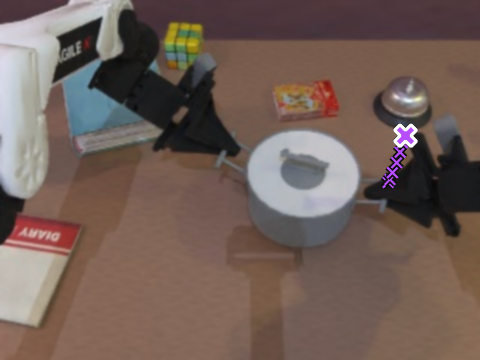} \\[1pt]
    \tasklab{Stack bowls}
     & \includegraphics[width=\qualpanelw]{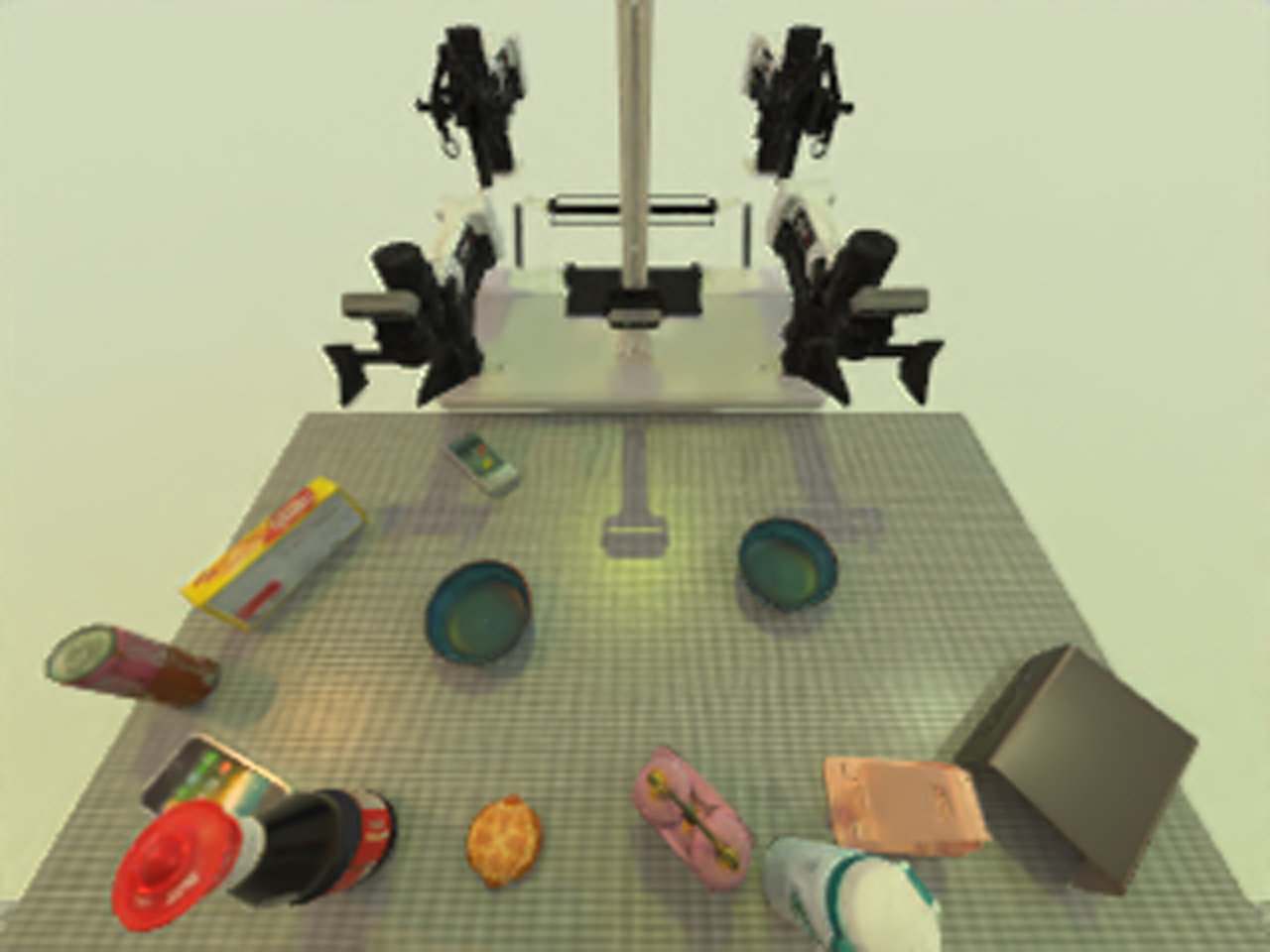}
     & \includegraphics[width=\qualpanelw]{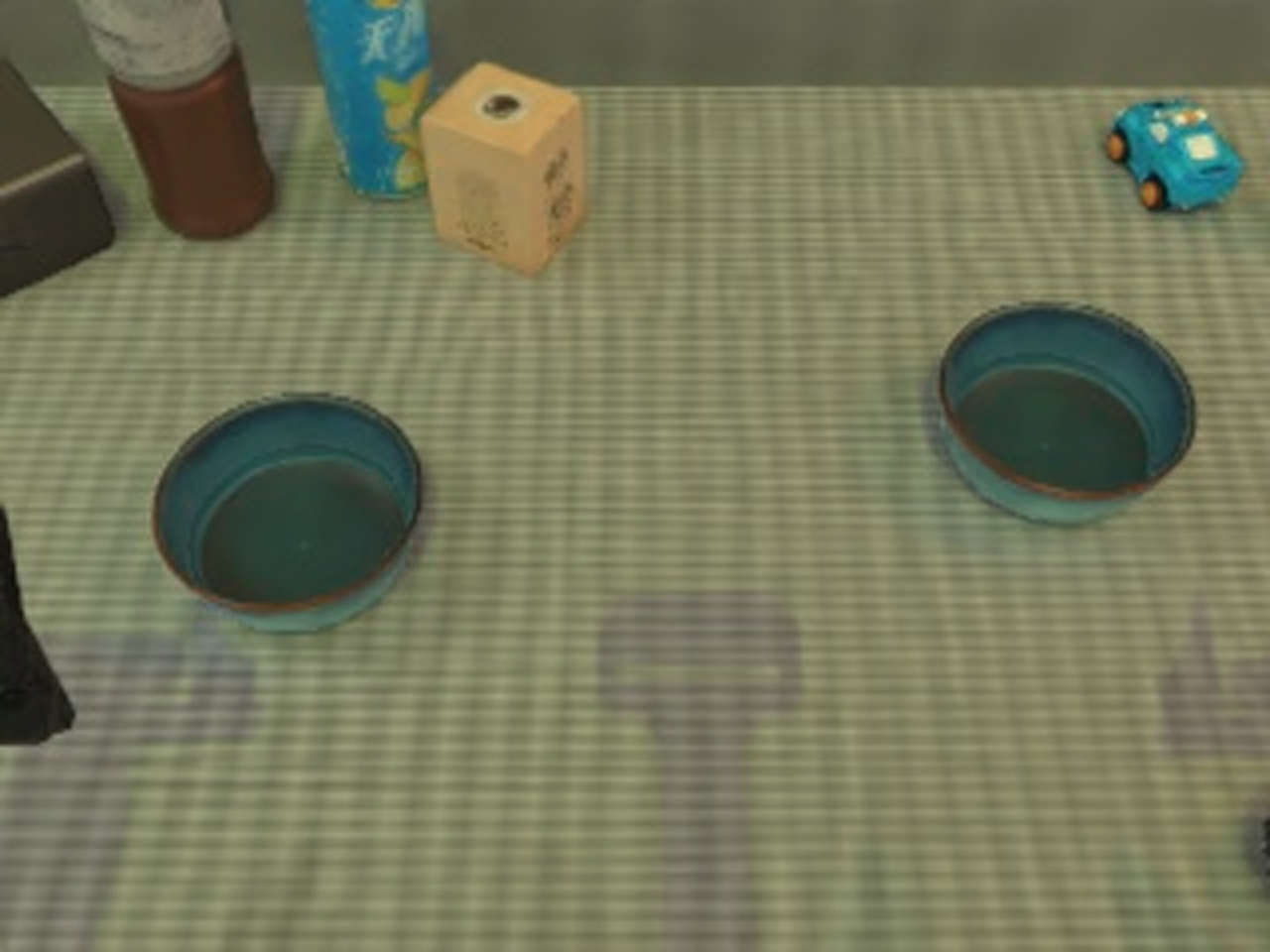}
     & \includegraphics[width=\qualpanelw]{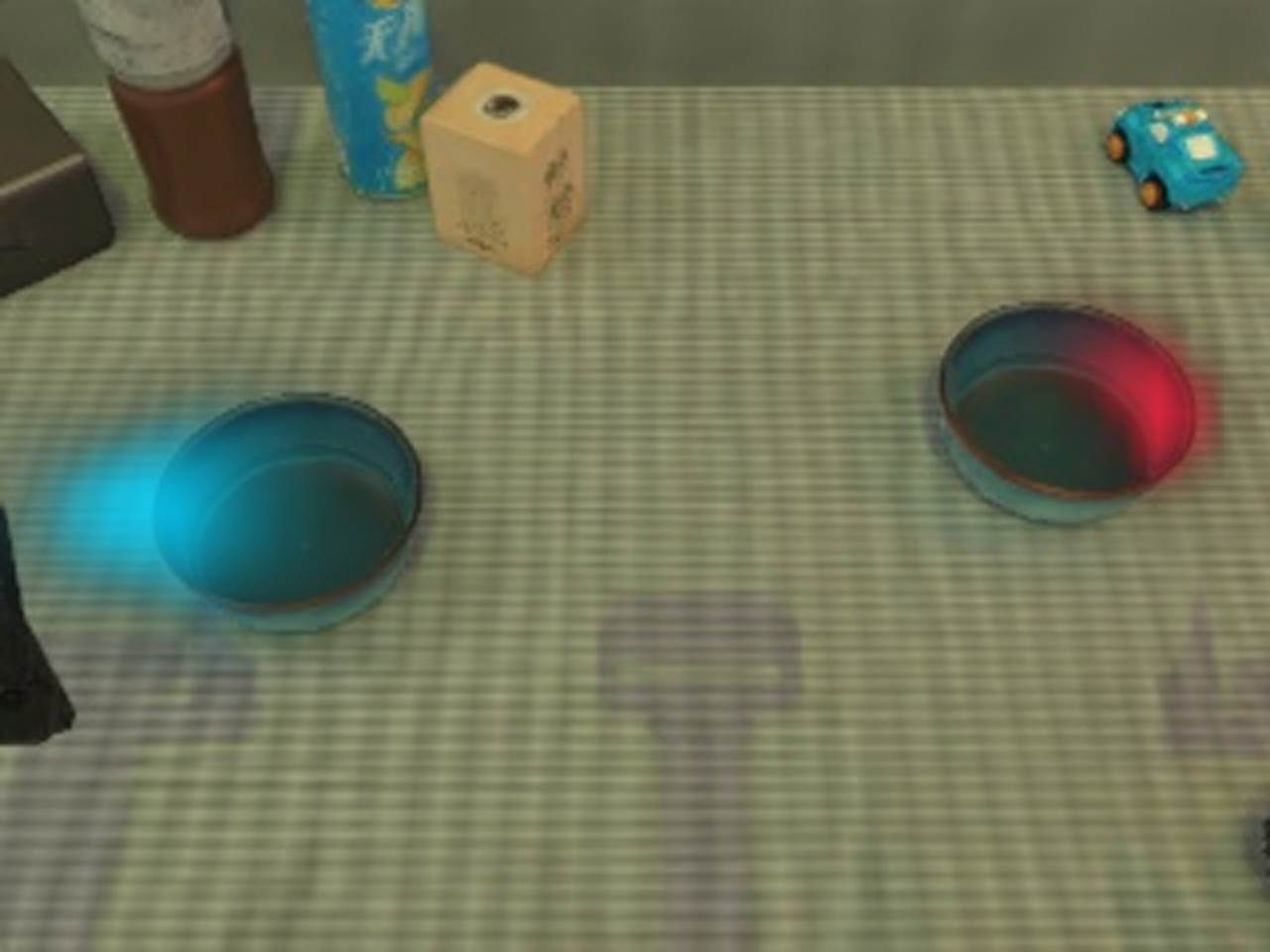}
     & \includegraphics[width=\qualpanelw]{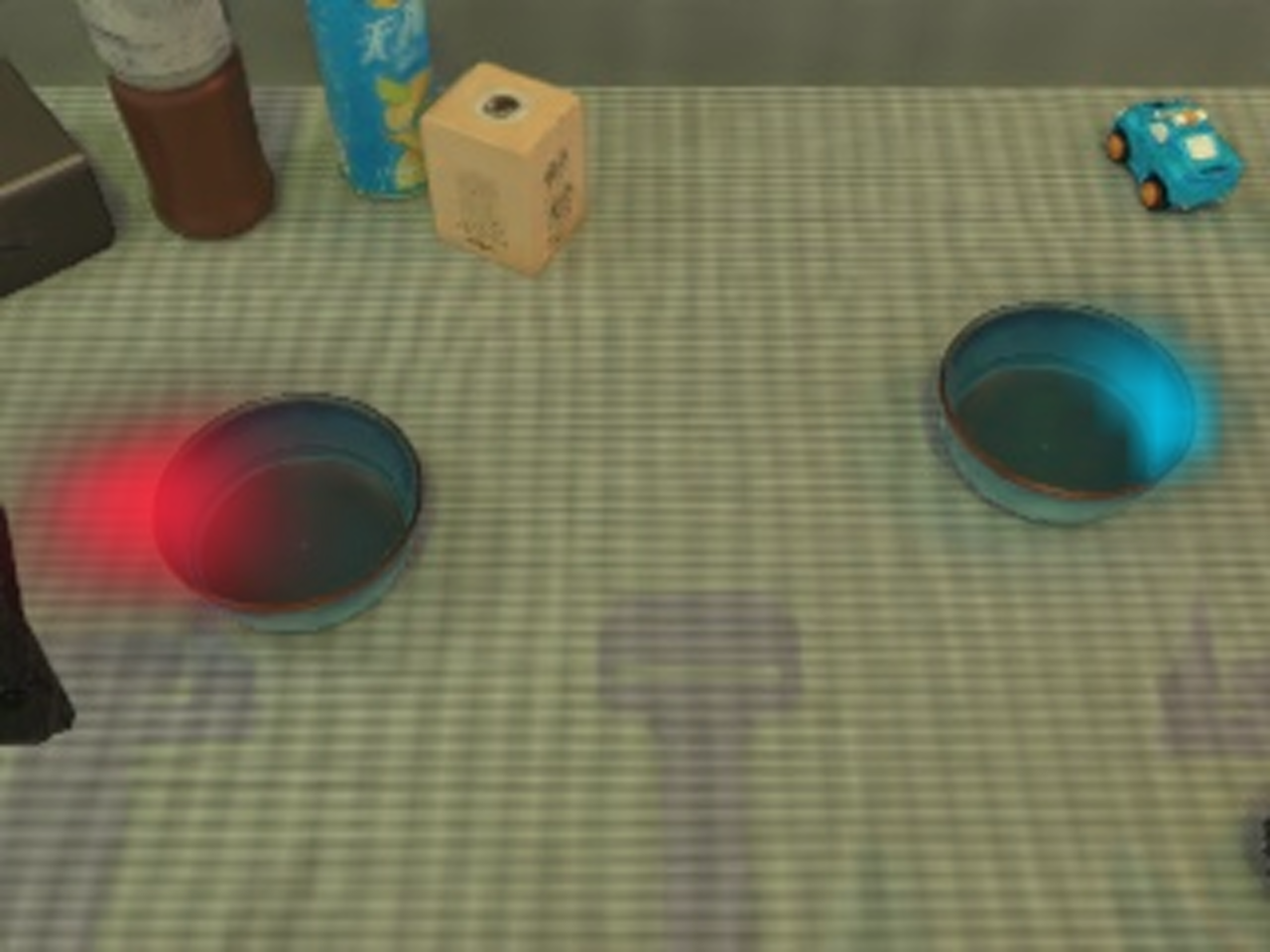}
     & \includegraphics[width=\qualpanelw]{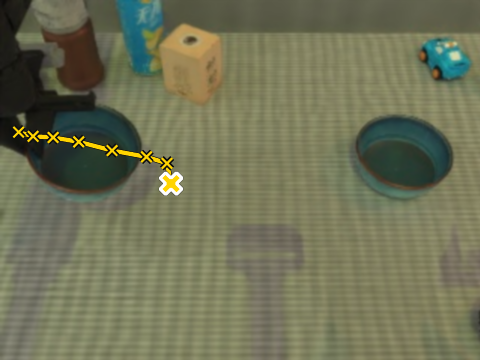}
     & \includegraphics[width=\qualpanelw]{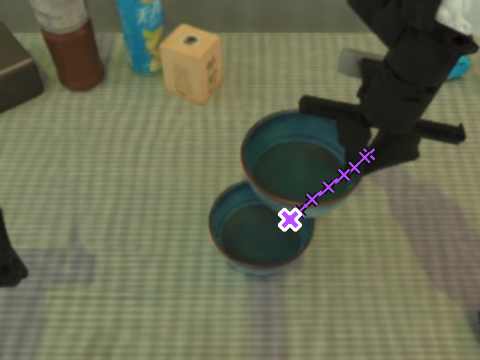} \\[1pt]
    \tasklab{Hang mug}
     & \includegraphics[width=\qualpanelw]{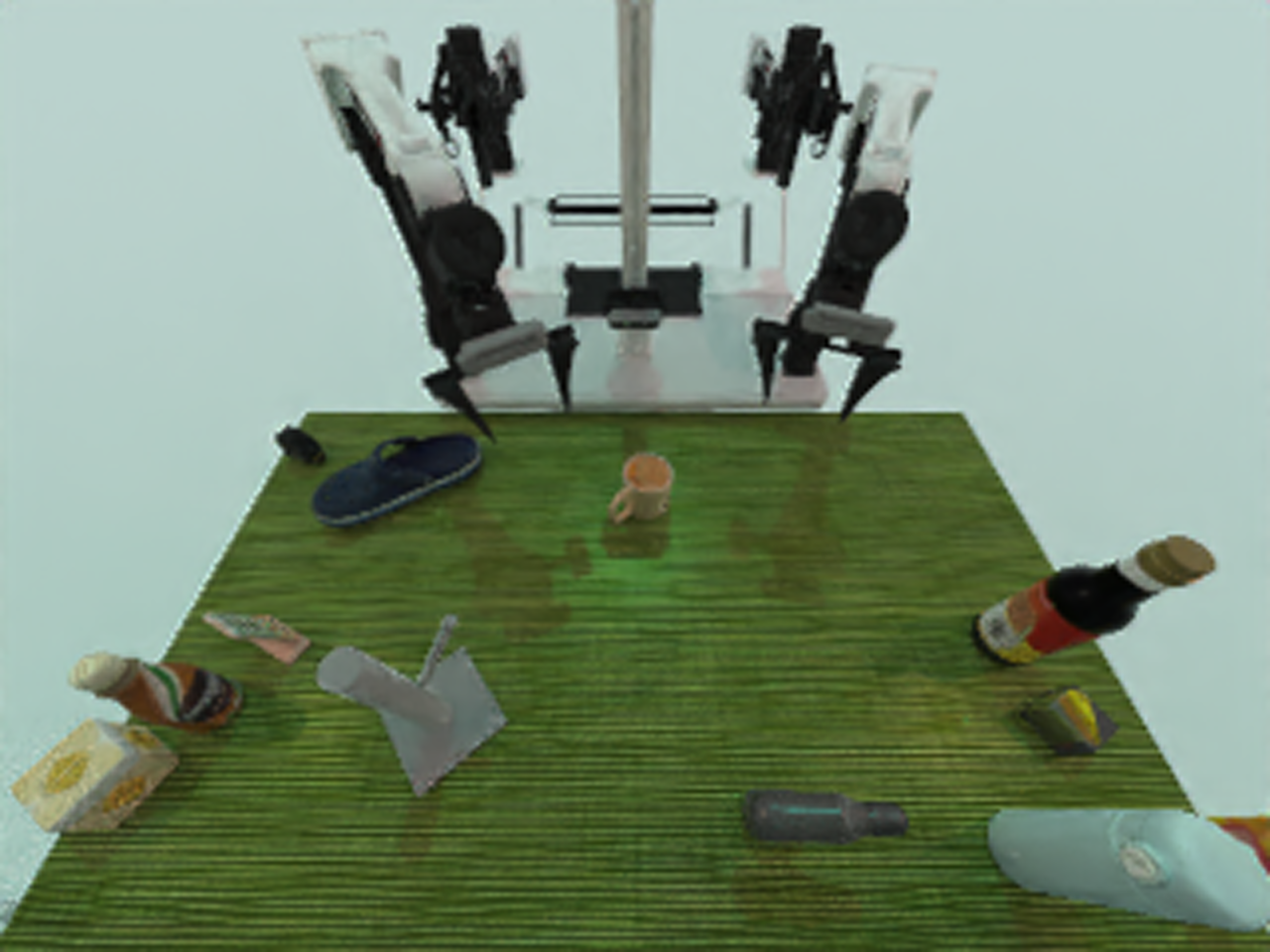}
     & \includegraphics[width=\qualpanelw]{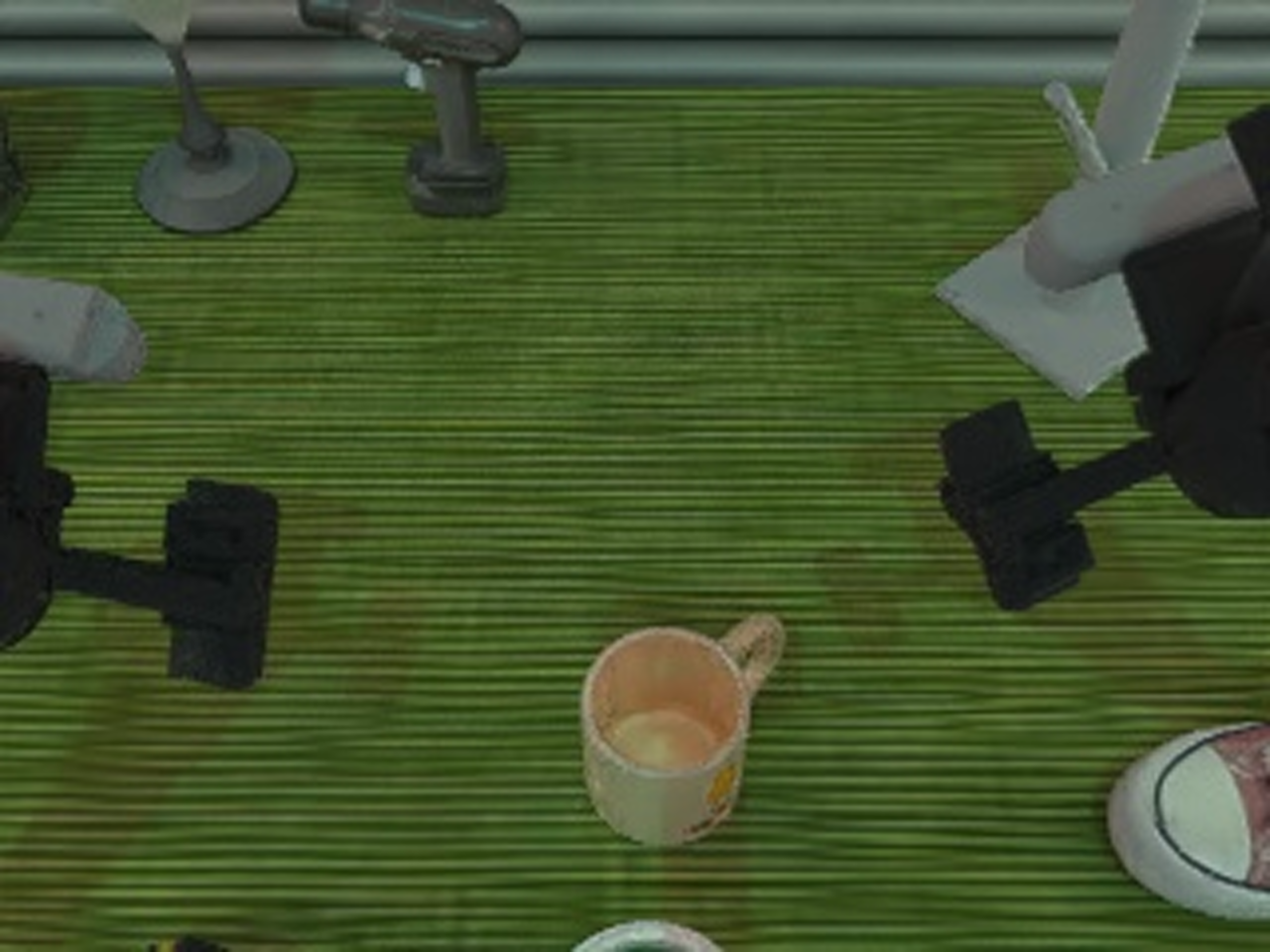}
     & \includegraphics[width=\qualpanelw]{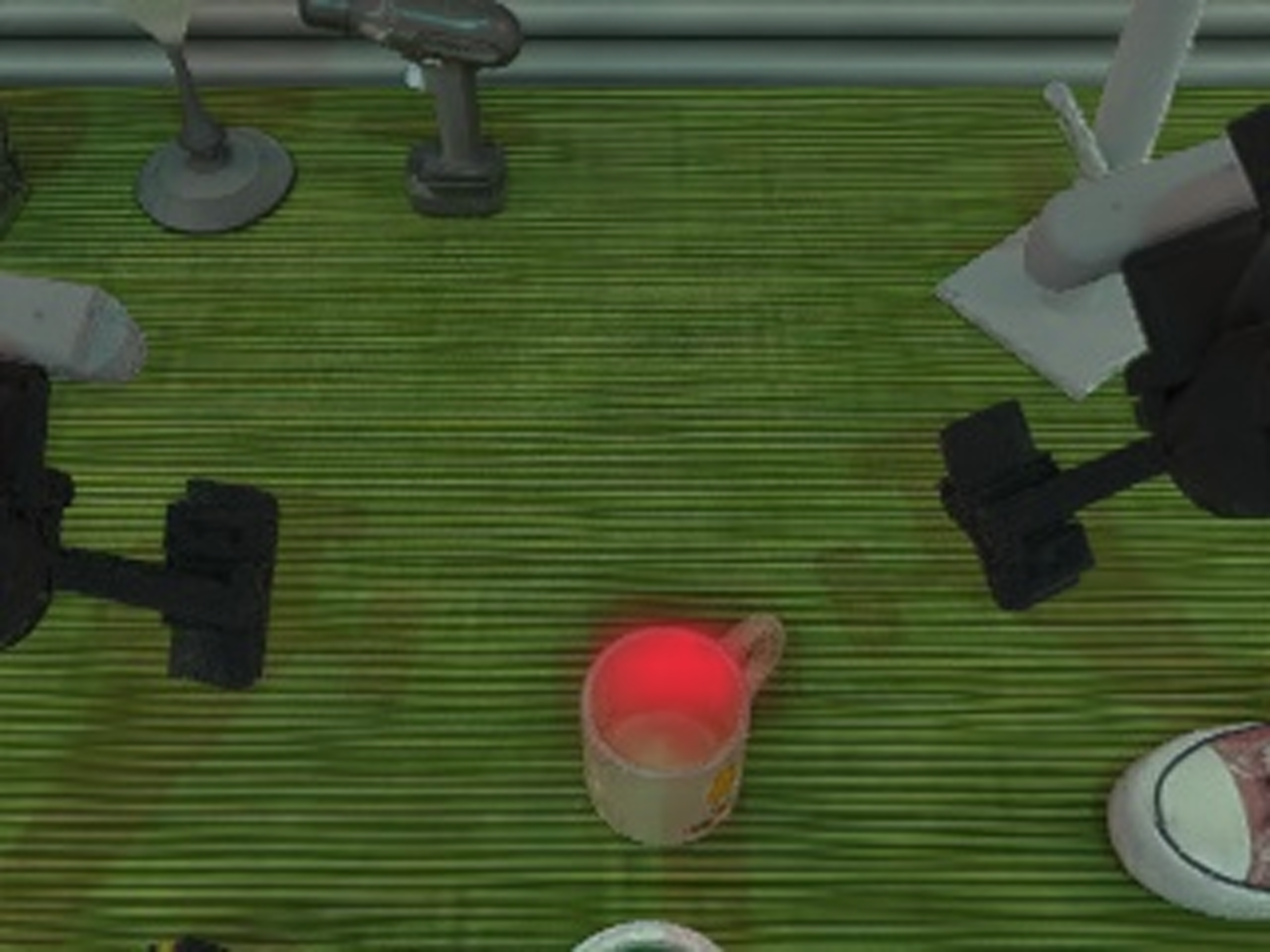}
     & \includegraphics[width=\qualpanelw]{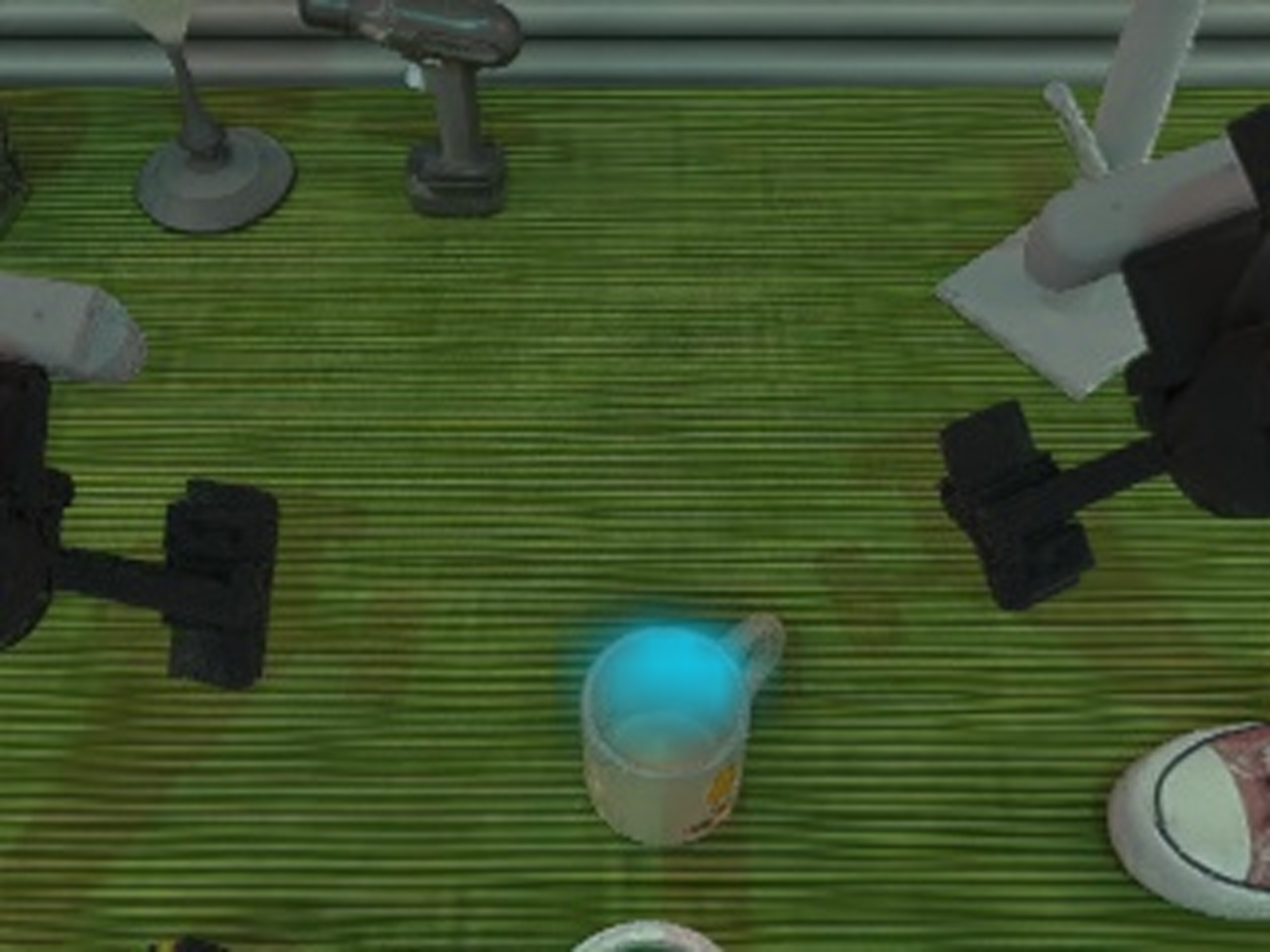}
     & \includegraphics[width=\qualpanelw]{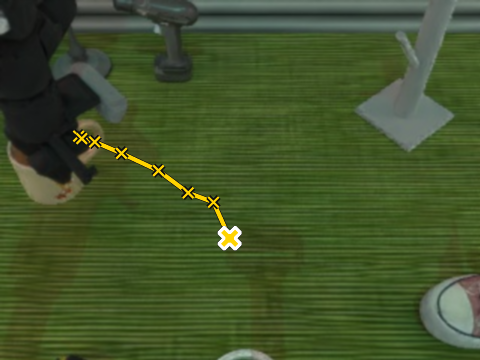}
     & \includegraphics[width=\qualpanelw]{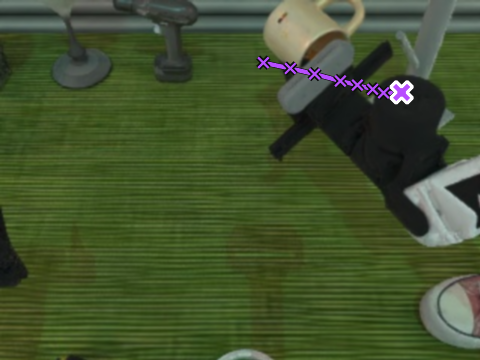} \\
  \end{tabular}
  \par\vspace{2pt}
  {\scriptsize Affordance: \textcolor{ownaff}{\rule{7pt}{5pt}} own\quad
   \textcolor{partneraff}{\rule{7pt}{5pt}} partner\qquad
   Trace: \textcolor{lefttrace}{\rule{7pt}{1.5pt}} left arm\quad
   \textcolor{righttrace}{\rule{7pt}{1.5pt}} right arm}
  \caption{\textbf{Simulation affordance grounding and phase-based traces.}
  Each row shows a third-person view, the head-camera RGB, and both agents' predicted affordances and traces in an OOD scene. Cyan and red denote own and partner affordances; yellow and violet denote the left and right agents' traces. Crosses mark phase points and the large cross marks the endpoint.}
  \label{fig:qualitative}
  \label{fig:skae}
  \label{fig:trace}
\end{figure*}

Affordance targets are contact-only: at the gripper's next physical contact event, the centroid of the contact points is projected into the top-camera image. A Gaussian centered at this pixel is multiplied by the mask of the contacted object part, which is obtained by prompting a pretrained segmentation foundation model with the contact pixels (Appendix).

Let $H^{i,\mathrm{own}}_t$ and $H^{i,\mathrm{partner}}_t$ be the corresponding target heatmaps. Following the Which2Act module in~\cite{yu2026affordancevla}, we additionally encode the image region under arm $i$'s own contact mask into a frozen target embedding $c^i_t$, which a train-only head $\hat{c}^i_t=g_{\mathrm{w2a}}(z^i_{\mathrm{aff},t})$ regresses. SKAE is trained with
\[
\mathcal{L}_{\mathrm{SKAE}}=\sum_{i\in\mathcal{I}}\Big[\sum_{\omega}\mathcal{L}_{\mathrm{hm}}(\hat{H}^{i,\omega}_t,H^{i,\omega}_t)+\lambda_{\mathrm{w2a}}\mathcal{L}^{i}_{\mathrm{w2a}}\Big]+\lambda_{\mathrm{asa}}\mathcal{L}_{\mathrm{asa}},
\]
where $\omega\in\{\mathrm{own},\mathrm{partner}\}$, the heatmap loss $\mathcal{L}_{\mathrm{hm}}$ fits each predicted heatmap to its target in value, region overlap, and peak location, $\mathcal{L}^{i}_{\mathrm{w2a}}$ is a cosine loss between $\hat{c}^i_t$ and $c^i_t$, $\mathcal{L}_{\mathrm{asa}}$ is a cross-entropy on the ASA score over frames where only one arm's affordance target is valid, and the $\lambda$ are loss weights. The trace module and the policies are trained afterwards on the frozen SKAE.

\subsection{3D Phase-Based Trace}\label{trace}

For each arm, the phase-trace module represents the intended motion of the gripper at the current phase as $K$ points $\tau^i_t=[p^i_{t,1},\ldots,p^i_{t,K}]$ with $p^i_{t,k}=(u^i_{t,k},v^i_{t,k},d^i_{t,k})$, where $(u,v)$ is a top-camera pixel coordinate and $d$ is the corresponding camera-frame depth, all normalized to $[0,1]$.

Expert trace labels are extracted from the demonstration dataset. For each frame $t$ and arm $i$, the current phase target time $t^{i}_{\mathrm{tar}}$ is first identified from a per-arm phase timeline built from end-effector motion and gripper events. The 3D tool-center positions $\bm{x}^i_{t_k}$ at $K$ temporal locations $t_k\in[t,t^{i}_{\mathrm{tar}}]$ are then projected as $p^i_{t,k}=\Pi(\bm{x}^i_{t_k})$, where $\Pi(\cdot)$ applies the calibrated camera intrinsics and extrinsics and returns the projected pixel together with its depth.

The trace predictor is trained as a supervised regression module with a separate conditioner per arm. The conditioner of arm $i$, with parameters $\eta$, computes
\[
\kappa^i_t=\mathcal{C}^{i}_{\eta}\big(E^i_t,\bm{r}^i_t,F^{hand,i}_t,f^{top}_t,f^{hand,i}_t,f^{\ell},\tilde{\bm{q}}^i_t,\tilde{\bm{q}}^{-i}_t\big),
\]
where $E^i_t$ are SKAE's affordance-supervised tokens and $\bm{r}^i_t=(\varrho(\hat{H}^{i,\mathrm{own}}_t),\hat{s}^i_t)$ is SKAE's decoded affordance readout, which pairs the own-affordance location $\varrho(\hat{H}^{i,\mathrm{own}}_t)$ with the ASA score. Grounding the trace on affordance-supervised tokens and readout lets it inherit SKAE's robustness to OOD scenes. Each point is decoded from $\kappa^i_t$ by a soft-argmax over predicted distributions on a pixel grid and over depth bins (Appendix), and we write $\hat{\tau}^i_t$ for the predicted trace. The module is trained with a masked mean-squared error $\mathcal{L}_{\mathrm{trace}}$ over the projected points. Fig.~\ref{fig:qualitative} shows both agents' predicted traces.

\subsection{Closed-Loop Execution}

During execution, the system repeatedly replans in closed loop. At each step, images are encoded by the frozen backbones, SKAE produces affordance latents and spatial tokens, the trace module provides camera-space guidance, and the two messages are exchanged. Each denoiser then samples an action chunk $\hat{a}^{i}_{t:t+T_a-1}$. Only the first $T_e<T_a$ actions are executed before the system receives a new observation and replans. Successive chunks are made consistent by real-time chunking~\cite{black2025rtc}, which guides each new chunk toward the unexecuted remainder of the previous one during denoising.

\section{Simulation Experiments}
\label{sec:sim}

We test on $15$ selected bimanual tasks of RoboTwin~2.0~\cite{chen2025robotwin2}. Each task trains on $550$ demonstrations ($50$ clean and $500$ randomized) and is evaluated on held-out OOD scenes with unseen object placements, lighting and backgrounds, $30$ rollouts ($15$ clean and $15$ randomized) per cell of Tables~\ref{tab:sr}--\ref{tab:fault_delay}. 

Besides the full policy we retrain five architecture ablations: \emph{no communication} ($\mathcal{M}=\emptyset$, Section~\ref{sec:comm}), \emph{centralized execution} (one actor over joint observation producing joint action chunk, i.e.\ CTCE), \emph{no partner affordance} (SKAE predicts only the agent's own affordance target, no partner prediction), \emph{no affordance} (DiT drops $z^i_{\mathrm{aff}}$ and trace conditions on ungrounded backbone tokens), and \emph{no trace}. We also compare our methods with the decentralized diffusion policy LatentToM (LToM)~\cite{he2025latenttom}, trained per task as ours, and two fine-tuned monolithic VLA baselines, $\pi_{0.5}$~\cite{intelligence2025pi05} and DM0.5~\cite{dm05}. All methods are evaluated on identical hardware-task configurations and share consistent hyperparameters.

\subsection{Task Success Rate}
\label{subsec:sr}

Table~\ref{tab:sr} reports full-task success with no fault injected. CoDDP averages $79\%$, level with $\pi_{0.5}$ ($80\%$) and slightly behind DM0.5 ($87\%$), while LToM, which does not target scene-level distribution shifts, reaches $12\%$ on these OOD scenes. Removing communication costs only $5$ points, since the shared top camera and SKAE's partner prediction still carry implicit coordination.
Centralized execution reaches $75\%$, so factorizing the policy does not degrade performance. The no-partner-affordance variant drops to $66\%$, since without partner supervision SKAE assigns roles less reliably.
Removing the affordance or the trace drops success to $53\%$ and $39\%$, mostly on precise placement tasks.

\begin{table*}[t]
  \centering
  \footnotesize
  \begin{tabular}{lcccccccccc}
    \toprule
    & \multicolumn{6}{c}{CoDDP} && \multicolumn{3}{c}{Baselines} \\
    \cmidrule(lr){2-7} \cmidrule(lr){9-11}
    Task & Ours & w/o comm. & Centr. & w/o partner aff. & w/o aff. & w/o trace & \vsep & LToM & $\pi_{0.5}$ & DM0.5 \\
    \cmidrule(lr){1-7} \cmidrule(lr){9-11}
    click alarmclock       & 97 & {\bc 100} & {\bc 100} & 97 & 67 & 43 & \vsep & 23 & {\bc 100} & {\bc 100} \\
    dump bin bigbin        & {\bc 100} & 97 & 93 & 93 & 90 & 87 & \vsep & 13 & 97 & 97 \\
    grab roller            & {\bc 100} & 87 & {\bc 100} & 97 & 90 & 70 & \vsep & 33 & {\bc 100} & {\bc 100} \\
    handover mic           & {\bc 100} & 90 & 93 & 87 & 77 & 80 & \vsep & 7 & 90 & {\bc 100} \\
    hanging mug            & 47 & 40 & 37 & 33 & 30 & 13 & \vsep & 3 & 43 & {\bc 63} \\
    lift pot               & {\bc 63} & 53 & {\bc 63} & 57 & 37 & 40 & \vsep & 10 & 60 & 57 \\
    pick diverse bottles   & 53 & 50 & 30 & 37 & 53 & 20 & \vsep & 30 & 50 & {\bc 77} \\
    place bread basket     & 83 & 73 & 70 & 67 & 23 & 20 & \vsep & 0 & 90 & {\bc 97} \\
    place bread skillet    & 77 & 70 & 63 & 47 & 40 & 13 & \vsep & 0 & {\bc 80} & 70 \\
    place burger fries     & 90 & 77 & 93 & 70 & 67 & 63 & \vsep & 3 & {\bc 100} & 93 \\
    place can basket       & 73 & 77 & 63 & 57 & 37 & 10 & \vsep & 27 & 70 & {\bc 87} \\
    place cans plasticbox  & 87 & 87 & 83 & 60 & 27 & 23 & \vsep & 10 & 93 & {\bc 97} \\
    place dual shoes       & 80 & 73 & 77 & 63 & 33 & 27 & \vsep & 0 & 70 & {\bc 97} \\
    scan object            & 43 & 40 & 53 & 30 & 37 & 7 & \vsep & 0 & 70 & {\bc 80} \\
    stack bowls two        & 97 & {\bc 100} & {\bc 100} & 90 & 83 & 67 & \vsep & 20 & 93 & 97 \\
    \cmidrule(lr){1-7} \cmidrule(lr){9-11}
    Average                & 79 & 74 & 75 & 66 & 53 & 39 & \vsep & 12 & 80 & {\bc 87} \\
    \bottomrule
  \end{tabular}
  \caption{\textbf{Full-Task Success Rates.} Evaluation on held-out OOD scenes with unseen object placements, lighting and backgrounds under no fault injection. Ours denotes the full CoDDP policy with communication.}
  \label{tab:sr}
\end{table*}

\subsection{Fault Isolation and Asynchronous Execution}
\label{subsec:fault}

Both properties follow from where the execution boundary lies, so we perturb one agent at serve time. The fault enters that agent's payload before preprocessing, while the shared top camera and instruction are not touched.  The choice of faulted agent is pooled across left and right arms. Tables~\ref{tab:fault_sensor} and~\ref{tab:fault_delay} report the two experiments.

For \emph{fault isolation} we corrupt one agent's proprioception and its own hand camera with identical Gaussian noise across methods. We report the intact arm's own subtask success rates.

CoDDP keeps the intact arm at $64\%$, well above the monoliths DM0.5 ($30\%$) and $\pi_{0.5}$ ($18\%$), whose joint conditioning passes the fault to both arms. The no-communication variant reaches $82\%$. This gap is the cost of the channel, which passes the corrupted state to the intact agent.

For \emph{asynchronous execution} we hold one arm's commanded action at its current pose for the first $40$ environment steps. Since both arms' states remain untouched, we report full-task success rates.

CoDDP is the best method at $75\%$, losing only $4$ points from its no-fault success against $26$ points for $\pi_{0.5}$ and $34$ points for DM0.5, whose joint action chunks assume arms moving in lockstep.

\begin{table}[t]
  \centering
  \footnotesize
  \setlength{\tabcolsep}{2.5pt}
  \begin{tabular}{lcccccc}
    \toprule
    & \multicolumn{2}{c}{CoDDP} && \multicolumn{3}{c}{Baselines} \\
    \cmidrule(lr){2-3} \cmidrule(lr){5-7}
    Task & Ours & w/o comm. & \vsepn & LToM & $\pi_{0.5}$ & DM0.5 \\
    \cmidrule(lr){1-3} \cmidrule(lr){5-7}
    click alarmclock       & 0 & {\bc 100} & \vsepn & 17 & {\bc 100} & 63 \\
    dump bin bigbin        & {\bc 100} & {\bc 100} & \vsepn & 13 & 0 & 30 \\
    grab roller            & 77 & {\bc 80} & \vsepn & 0 & 10 & 20 \\
    handover mic           & {\bc 100} & {\bc 100} & \vsepn & 3 & 0 & 60 \\
    hanging mug            & 87 & {\bc 100} & \vsepn & 43 & 0 & 50 \\
    lift pot               & {\bc 83} & 63 & \vsepn & 13 & 0 & 0 \\
    pick diverse bottles   & {\bc 63} & 60 & \vsepn & 37 & 17 & 47 \\
    place bread basket     & 60 & {\bc 67} & \vsepn & 0 & 13 & 0 \\
    place bread skillet    & {\bc 83} & 80 & \vsepn & 13 & 0 & 50 \\
    place burger fries     & 60 & {\bc 87} & \vsepn & 27 & 13 & 0 \\
    place can basket       & 73 & {\bc 77} & \vsepn & 7 & 0 & 3 \\
    place cans plasticbox  & 17 & {\bc 90} & \vsepn & 27 & 50 & 0 \\
    place dual shoes       & 73 & {\bc 83} & \vsepn & 17 & 0 & 37 \\
    scan object            & {\bc 73} & {\bc 73} & \vsepn & 7 & 13 & 17 \\
    stack bowls two        & 13 & 63 & \vsepn & 37 & 50 & {\bc 70} \\
    \cmidrule(lr){1-3} \cmidrule(lr){5-7}
    Average                & 64 & {\bc 82} & \vsepn & 17 & 18 & 30 \\
    \bottomrule
  \end{tabular}
  \caption{\textbf{Fault Isolation.} At serve time, one agent's proprioception receives Gaussian noise and its own hand camera is replaced by random pixels. We report the intact arm's own subtask success rates.}
  \label{tab:fault_sensor}
\end{table}

\begin{table}[t]
  \centering
  \footnotesize
  \setlength{\tabcolsep}{2.5pt}
  \begin{tabular}{lcccccc}
    \toprule
    & \multicolumn{2}{c}{CoDDP} && \multicolumn{3}{c}{Baselines} \\
    \cmidrule(lr){2-3} \cmidrule(lr){5-7}
    Task & Ours & w/o comm. & \vsepn & LToM & $\pi_{0.5}$ & DM0.5 \\
    \cmidrule(lr){1-3} \cmidrule(lr){5-7}
    click alarmclock       & 97 & {\bc 100} & \vsepn & 23 & {\bc 100} & {\bc 100} \\
    dump bin bigbin        & {\bc 100} & {\bc 100} & \vsepn & 10 & {\bc 100} & 83 \\
    grab roller            & {\bc 97} & 83 & \vsepn & 17 & 73 & 83 \\
    handover mic           & {\bc 100} & 93 & \vsepn & 3 & {\bc 100} & {\bc 100} \\
    hanging mug            & 47 & 37 & \vsepn & 7 & 37 & {\bc 63} \\
    lift pot               & 70 & 37 & \vsepn & 7 & 3 & {\bc 83} \\
    pick diverse bottles   & 47 & {\bc 67} & \vsepn & 7 & 0 & 0 \\
    place bread basket     & {\bc 73} & 60 & \vsepn & 0 & 43 & 37 \\
    place bread skillet    & {\bc 63} & 53 & \vsepn & 0 & {\bc 63} & 23 \\
    place burger fries     & {\bc 90} & 73 & \vsepn & 3 & 67 & 10 \\
    place can basket       & 73 & {\bc 80} & \vsepn & 27 & 70 & {\bc 80} \\
    place cans plasticbox  & 77 & {\bc 80} & \vsepn & 7 & 20 & 0 \\
    place dual shoes       & {\bc 67} & {\bc 67} & \vsepn & 0 & 17 & 0 \\
    scan object            & 23 & {\bc 33} & \vsepn & 0 & 27 & 30 \\
    stack bowls two        & {\bc 100} & {\bc 100} & \vsepn & 27 & 93 & 97 \\
    \cmidrule(lr){1-3} \cmidrule(lr){5-7}
    Average                & {\bc 75} & 71 & \vsepn & 9 & 54 & 53 \\
    \bottomrule
  \end{tabular}
  \caption{\textbf{Asynchronous Execution.} One arm's commanded action is held at its current pose for the first $40$ environment steps. We report full-task success rates.}
  \label{tab:fault_delay}
\end{table}

\subsection{Value of Communication}
\label{subsec:voc}

We quantify how much the received partner message helps generate the agent's own action across tasks, given its local inputs $o^i$. Keeping these inputs fixed, we sample actions with messages drawn uniformly from four recorded candidates and estimate the Barber–Agakov lower bound~\cite{barber2003im,poole2019variational},
\begin{equation*}
I(a^i;m^{-i}\mid o^i) \geq
\mathbb{E}\!\left[\log\frac{q_{\phi}(m^{-i}\mid a^i,o^i)}{p(m^{-i}\mid o^i)}\right],
\end{equation*}
where $p(m^{-i}\mid o^i)$ is the known message-selection prior and $q_{\phi}$ is a learned categorical posterior evaluated on held-out action samples. We average the estimates over both arms and scenes.

The no-communication variant receives no messages, so its channel information is zero.

\begin{figure}[t]
  \centering
  \includegraphics[width=\columnwidth]{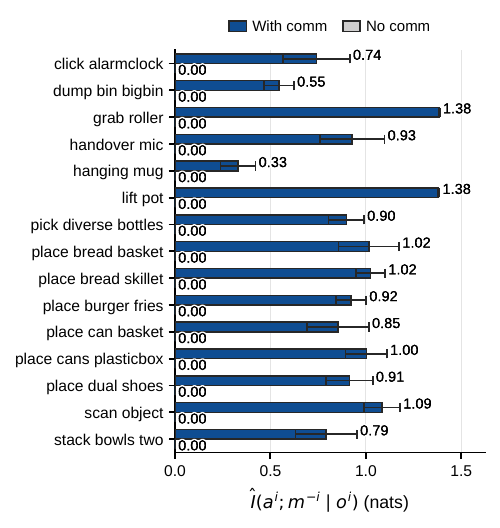}
  \caption{\textbf{Value of Communication.} Estimated information lower bounds (nats). Bars show means and error bars show standard deviations across tasks.}
  \label{fig:voc}
\end{figure}

Fig.~\ref{fig:voc} reports the resulting estimates in nats, the unit of information given by the natural logarithm; one nat equals $1/\ln 2\approx 1.44$ bits. Because the message is one of four equally likely candidates, the estimate cannot exceed the prior entropy $\ln 4\approx 1.39$ nats ($2$ bits), which it reaches only when the action identifies the received message exactly. CoDDP averages $0.92$ nats ($1.33$ bits), so each agent's action depends on the message it receives. Estimates saturate on grab roller and lift pot, where the arms hold one object together, and are lowest on hanging mug and dump bin bigbin, where they act mostly in turn.

\section{Conclusion}
\label{sec:conclusion}

We presented CoDDP, which factorizes a bimanual policy into two diffusion policies that are trained jointly and exchange a compact message at execution. Own and partner affordances and the phase-based trace give each agent the coordination cues that a monolithic policy obtains from the joint observation. As a result, decentralized execution is comparable with fine-tuned monolithic VLAs in task success, and more robust than them when one arm's proprioception is corrupted or its motion is delayed. The message carries measurable information about each agent's action, but it also passes a corrupted partner state to the intact agent.

\emph{Limitations.} Limited compute restricted each agent's policy to a task-specific diffusion model on frozen backbones rather than a fine-tuned VLA. Both agents share the top camera, so even the no-communication variant remains coupled through that image, and a fault in it would reach both agents. Scaling to more agents is also limited due to constraints like data collection.

\emph{Future work.} We will replace each agent's diffusion policy with a VLA fine-tuned on top of the SKAE and trace modules. Because the policy, affordances, and message are all defined per agent, the formulation also extends in principle to more than two agents.

\appendix[Method Details]
\label{app:method}

\emph{Backbones and SKAE.} The vision encoder is DINOv3~\cite{simeoni2025dinov3}, the vision-language encoder is SigLIP~2~\cite{tschannen2025siglip2}, whose global image embedding of the top camera is grouped with $f^{top}_t$, and the SKAE encoder is a transformer encoder. The part masks of the affordance targets are obtained by prompting SAM~2~\cite{ravi2024sam2} with the contact pixels. Contact events and contact points are read directly from the simulator.
For a predicted heatmap $\hat{H}$ and its target $H$, the heatmap loss $\mathcal{L}_{\mathrm{hm}}=\mathcal{L}_{\mathrm{bce}}+\lambda_{\mathrm{dice}}\mathcal{L}_{\mathrm{dice}}+\lambda_{\mathrm{coord}}\mathcal{L}_{\mathrm{coord}}$ combines a peak-weighted cross-entropy, a soft-Dice term, and a coordinate term with weights $\lambda_{\mathrm{dice}}$ and $\lambda_{\mathrm{coord}}$. Let $\mathrm{BCE}(\cdot,\cdot)$ be the pixel-wise binary cross-entropy and $\langle\cdot,\cdot\rangle$ the pixel average of the element-wise product. The cross-entropy $\mathcal{L}_{\mathrm{bce}}(\hat{H},H)=\langle 1+\lambda_{\mathrm{pk}}H,\mathrm{BCE}(\hat{H},H)\rangle+\lambda_{\mathrm{inv}}\mathcal{B}(\hat{H})$ fits every pixel value, with $\lambda_{\mathrm{pk}}$ up-weighting pixels inside the target so that the small affordance region is not outweighed by the background, while $\mathcal{B}(\hat{H})=\langle 1,\mathrm{BCE}(\hat{H},0)\rangle$, weighted by $\lambda_{\mathrm{inv}}$, suppresses $\hat{H}$ on invalid frames, where the arm has no affordance target. The soft-Dice term $\mathcal{L}_{\mathrm{dice}}(\hat{H},H)=1-2\langle\hat{H},H\rangle/(\langle\hat{H},\hat{H}\rangle+\langle H,H\rangle)$ measures region overlap independently of the region's area, so the prediction covers the contacted part not just its peak. The coordinate term $\mathcal{L}_{\mathrm{coord}}(\hat{H},H)=\|\sigma(\hat{H})-\sigma(H)\|_2^2$ aligns the predicted and target locations, where the spatial soft-argmax $\sigma(\cdot)$ returns the normalized centroid of a heatmap. All terms except $\mathcal{B}$ are averaged over valid frames.
The Which2Act target $c^i_t$ is the DINOv3 embedding of the image crop that bounds arm $i$'s own contact mask, and $\mathcal{L}^{i}_{\mathrm{w2a}}=1-\cos(\hat{c}^i_t,c^i_t)$ over valid frames. The ASA loss is $\mathcal{L}_{\mathrm{asa}}=-\log\hat{s}^{i^\star}_t$, where $i^\star$ is the only arm whose affordance target is valid. The heatmap terms train the decoder and, through $E^i_t$, the encoder, while the w2a and ASA terms reach the encoder through $z^i_{\mathrm{aff},t}$.

\emph{Phase-based trace.} In the conditioner $\mathcal{C}^{i}_{\eta}$, $E^i_t$ and $F^{hand,i}_t$ are aggregated by learned-query attention pooling, and the location $\varrho(\cdot)$ in the readout $\bm{r}^i_t$ is a softmax-weighted centroid over the heatmap's strongest cells. Each point is decoded from $\kappa^i_t$ by a soft-argmax over a predicted distribution, $(\hat{u}^i_{t,k},\hat{v}^i_{t,k})=\sum_{\bm{g}}\mathrm{softmax}(\zeta^i_{t,k}/\vartheta)_{\bm{g}}\,\bm{g}$ and $\hat{d}^i_{t,k}=\sum_{\beta=1}^{B}\mathrm{softmax}(\delta^i_{t,k}/\vartheta_{d})_{\beta}\,\bar{d}_{\beta}$, where $\zeta^i_{t,k}$ and $\delta^i_{t,k}$ are logits predicted from $\kappa^i_t$, $\bm{g}$ ranges over normalized grid coordinates, $\bar{d}_{\beta}$ is the center of the $\beta$-th of $B$ normalized depth bins, and $\vartheta,\vartheta_{d}$ are learned temperatures.
The training loss is a masked mean-squared error over the projected points, $\mathcal{L}_{\mathrm{trace}}=\sum_{i\in\mathcal{I}}\mathcal{L}^{i}_{\mathrm{pt}}$ with $\mathcal{L}^{i}_{\mathrm{pt}}=\sum_{k=1}^{K}\mu^i_{t,k}\|\hat{p}^{i}_{t,k}-p^{i}_{t,k}\|_2^2/\max(1,\sum_{k=1}^{K}\mu^i_{t,k})$, where $\hat{p}^{i}_{t,k}=(\hat{u}^i_{t,k},\hat{v}^i_{t,k},\hat{d}^i_{t,k})$ and $\mu^i_{t,k}\in\{0,1\}$ indicates if the projected point is valid; the maximum keeps the loss defined when no point is valid.

\emph{Policy conditioning.} Each trace embedding $\mathrm{Emb}^{i}$ is a sinusoidal feature expansion followed by an MLP, which lifts the low-dimensional trace so that it is not diluted by the visual condition. During training the condition is replaced by a learned unconditional embedding with probability $p_{\mathrm{cfg}}$, so classifier-free guidance~\cite{ho2022cfg} is available at sampling time.

\bibliographystyle{IEEEtran}
\bibliography{refs}

\end{document}